\documentclass[10pt,letterpaper]{article}
\usepackage{amsmath,amssymb,booktabs,graphicx}
\usepackage{algorithm,algpseudocode,placeins}
\usepackage[hidelinks]{hyperref}
\usepackage[textwidth=5.5in,textheight=9in,centering,headheight=12pt,headsep=16pt]{geometry}
\usepackage[no-math]{fontspec}
\usepackage{xcolor}
\definecolor{obrpurple}{HTML}{8064A2}
\definecolor{obrred}{HTML}{B84A53}
\definecolor{obrgray}{HTML}{707070}
\makeatletter
\renewcommand\normalsize{\@setfontsize\normalsize{10}{11}%
 \abovedisplayskip 7pt plus 2pt minus 2pt
 \belowdisplayskip \abovedisplayskip
 \abovedisplayshortskip 4pt plus 2pt
 \belowdisplayshortskip 5pt plus 2pt minus 2pt}
\normalsize
\renewcommand\section{\@startsection{section}{1}{\z@}%
 {-14pt plus -2pt minus -1pt}{6pt}%
 {\normalfont\fontsize{12}{14}\selectfont\bfseries}}
\renewcommand\subsection{\@startsection{subsection}{2}{\z@}%
 {-10pt plus -2pt minus -1pt}{4pt}%
 {\normalfont\fontsize{10}{11}\selectfont\bfseries}}
\renewcommand\subsubsection{\@startsection{subsubsection}{3}{\z@}%
 {-8pt plus -1pt minus -1pt}{3pt}%
 {\normalfont\fontsize{10}{11}\selectfont\itshape}}
\renewcommand\paragraph{\@startsection{paragraph}{4}{\z@}%
 {5pt plus 1pt minus 1pt}{-.6em}%
 {\normalfont\normalsize\bfseries}}
\long\def\@makecaption#1#2{%
 \vskip 5pt
 {\fontsize{9}{10.5}\selectfont\noindent #1: #2\par}%
 \vskip 4pt}
\renewcommand\maketitle{%
 \vspace*{8pt}
 {\raggedright\fontsize{16}{18}\selectfont\bfseries\@title\par}%
 \vspace{7pt}
 {\raggedright\normalfont\fontsize{11}{13}\selectfont\@author\par}%
 \vspace{5pt}{\small\@date\par}\vspace{8pt}}
\def\ps@preprint{%
 \def\@oddhead{\vbox{\hbox to\textwidth{\normalfont\fontsize{10}{11}\selectfont Preprint\hfil}\vskip 3pt\hrule height .3pt}}%
 \let\@evenhead\@oddhead
 \def\@oddfoot{\hfil\normalfont\fontsize{10}{11}\selectfont\thepage\hfil}%
 \let\@evenfoot\@oddfoot}
\makeatother
\renewenvironment{abstract}{%
 \begin{list}{}{\setlength{\leftmargin}{.36in}\setlength{\rightmargin}{.36in}}\item[]
 \begin{center}\normalsize\textbf{Abstract}\end{center}\vspace{-4pt}\normalsize}%
 {\end{list}\vspace{5pt}}

\newcommand{\Q}{\mathcal Q}
\newcommand{\U}{\mathcal U}
\newcommand{\Piv}{\mathcal P}
\newcommand{\Rfp}{\mathcal R_{\mathrm{FP8}}}
\title{\bfseries Low-Bit GDN Recurrent-State Quantization:\\Global Gradient-aware Allocation and Pure-Write Sliding Windows}
\author{%
 Qinghang Huang\quad Xuchu Huang\quad Xunzhao Yin\textsuperscript{*}\\[4pt]
 {\fontsize{10}{12}\selectfont Zhejiang University}\\[3pt]
 {\fontsize{9}{11}\selectfont \href{mailto:22560023@zju.edu.cn}{22560023@zju.edu.cn}\quad\href{mailto:xchuang_@zju.edu.cn}{xchuang\_@zju.edu.cn}\quad\href{mailto:xzyin1@zju.edu.cn}{xzyin1@zju.edu.cn}}\\[2pt]
 {\fontsize{9}{11}\selectfont\textsuperscript{*}Corresponding author.}}
\hypersetup{pdfauthor={Qinghang Huang, Xuchu Huang, Xunzhao Yin}}
\date{October 18, 2026}
\begin{document}
\maketitle
\begin{abstract}
Linear attention stores historical information in a recurrent state of fixed size, providing a compact memory representation for long-context inference. Reducing state precision further lowers per-request storage requirements, but quantized writeback errors enter subsequent recurrences, degrade prediction quality on long-reasoning tasks, and can be accompanied by repetitive generation and excessively long outputs. We propose Global Gradient-aware Allocation (GGA) and the Pure-Write Sliding Window (PW) to address precision allocation and recent-information maintenance. GGA combines candidate quantization residuals with gradient sensitivity to future prediction loss to allocate head bit widths under a model-wide mixed-precision budget. PW independently maintains recent pure writes as FP8 vector pairs, propagates them through the full GDN state transition, and merges them into the low-bit historical state only after they leave the window. Across seven evaluations on Qwen3.5-4B/9B, GGA allocation alone improves over STEPQuant@$4$ by $19.48$/$29.84$ percentage points. Adding PW-S8 yields seven-task macro-averages of $69.83\%$/$78.63\%$, improving over STEPQuant@$4$ by $38.39$/$41.21$ percentage points, respectively. Both precision reallocation and window maintenance improve task scores while mitigating excessively long outputs: at @$6$, GGA reduces mean generation length, averaged equally across tasks, by $10.45\%$/$14.65\%$ relative to STEPQuant; adding PW-S8 to GGA@$4$ further reduces mean generation length by $32.32\%$/$36.77\%$. Comparisons at similar storage budgets further show that using additional capacity to maintain recent writes yields higher task scores and shorter mean outputs than solely increasing the bit width of the accumulated state.
\end{abstract}

\section{Introduction}
\label{sec:introduction}
\suppressfloats[t]

Autoregressive inference requires the continuous retention of historical information. In standard attention, the KV cache grows with sequence length, whereas linear attention compresses historical interactions into a recurrent state of fixed size~\cite{linear-attention}. Architectures such as GLA, Gated DeltaNet (GDN), and Mamba-2 further improve the expressiveness and computational efficiency of recurrent memory~\cite{gla,gated-deltanet,mamba2}. Nevertheless, even a fixed-size state must be stored separately for each concurrent request and read and updated at every decoding step; its storage and memory-access costs therefore motivate compression in inference deployment~\cite{damp,stepquant,leapquant}.

Recurrent-state quantization faces both precision-allocation and error-feedback challenges. Perturbations to different heads have different effects on subsequent predictions, while writeback errors introduced by stepwise quantization enter future recurrences. DAMP protects high-risk channels using decay-weighted risk, STEPQuant jointly models error persistence and readout impact, and LeapQuant delays state quantization through window buffering~\cite{damp,stepquant,leapquant}. These studies provide an effective foundation for state compression and motivate two further questions: how final prediction loss can guide head precision allocation, and how recent information can be maintained according to write age.

For precision allocation, local candidate distortion and error lifetime alone do not determine changes in final predictions. State residuals of equal magnitude can have different effects because of their directions, subsequent query readouts, and downstream network mappings. We retain STEPQuant's candidate quantization formats and nominal budget, and weight candidate residuals by gradients of future language-model loss with respect to the writeback state to construct an additive risk proxy for Global Gradient-aware Allocation (GGA). For temporal maintenance, we separate pure writes in the GDN update into individual floating-point vector branches, apply the full state transition to them, and evict them individually into the low-bit historical state at a fixed age, yielding the Pure-Write Sliding Window (PW). This rolling maintenance differs from quantizing the entire state at the end of a window: the old state is still quantized at every step.

Our contributions are as follows:
\begin{itemize}
 \item We propose Global Gradient-aware Allocation (GGA) guided by future prediction loss. Along calibration trajectories, we estimate the sensitivity of future loss to the state at each writeback position and construct comparable head risks by weighting squared candidate quantization residuals coordinatewise with squared gradients. Retaining STEPQuant's candidate quantization formats and nominal budget, we combine FP16 pivot selection with discrete bit-width optimization to reallocate precision across layers and heads; gradient computation is used only for offline calibration.
 \item We propose the Pure-Write Sliding Window (PW). We separate recent pure writes from the GDN recurrence, represent each branch as a vector pair, apply the full state transition at subsequent steps, and merge branches into the low-bit historical state on a rolling basis according to write age. This design retains stepwise quantization of the old state while delaying the first incorporation of recent branches into the low-bit matrix. We provide a proof of recurrence equivalence under ideal conditions and an online FP8 maintenance algorithm with per-vector scales.

\end{itemize}

\section{Background and Related Work}
\label{sec:related-work}
\label{sec:preliminaries}
\suppressfloats[t]

\subsection{Linear Recurrent Models and State Compression}

Linear attention accumulates key/value interactions in a fixed-size state, enabling autoregressive inference through recurrent updates~\cite{linear-attention}. GLA adds data-dependent forgetting gates~\cite{gla}; Gated DeltaNet (GDN) combines gated forgetting with the delta rule to modify memory along the current key direction~\cite{gated-deltanet}; Mamba-2 connects selective state-space models and attention through structured state-space duality~\cite{mamba2}. Although these architectures avoid a state that grows with sequence length, each concurrent request retains its own state, which is read and updated at every decoding step. Compressing that state therefore remains relevant to inference storage.

State compression must account for how a memory representation is used. KIVI quantizes attention keys per channel and values per token, while retaining a floating-point buffer of recent KV entries~\cite{kivi}. For selective state-space models, Quamba2 groups recurrent-operator inputs according to persistent channel magnitudes and applies the grouping to an 8-bit state cache~\cite{quamba2}; Q-Mamba uses decoupled scaling along the state and channel axes and simulates quantized recurrences during blockwise reconstruction~\cite{qmamba}. In contrast to KV entries appended to a cache, recurrent states repeatedly become inputs to later updates, allowing writeback errors to propagate through subsequent computation.

\subsection{GDN Recurrence and Quantized Writeback}
\label{sec:recurrence}

Consider one state/value head in GDN, omitting layer and head indices. At update $t$, $S_{t-1}\in\mathbb R^{d_k\times d_v}$ is the incoming historical state and $S_t$ is the updated state; $S_0$ is the state at the end of prefill. The query and key are $q_t,k_t\in\mathbb R^{d_k}$, and the value is $v_t\in\mathbb R^{d_v}$. These are the processed activations entering the recurrent operator, including the model's normalization, scaling, and mapping from shared queries/keys to state heads. With retention coefficient $\gamma_t$ and key-update strength $\beta_t$, GDN updates the state and then reads the current head output:
\begin{equation}
 \begin{aligned}
 S_t&=\gamma_tS_{t-1}
 +\beta_tk_t\bigl(v_t-\gamma_tS_{t-1}^{\top}k_t\bigr)^{\top},\\
 o_t&=S_t^{\top}q_t.
 \end{aligned}
 \label{eq:gdn}
\end{equation}
The raw readout $o_t$ is followed by normalization, output gating, and projection in the downstream network. We separate the update into a state transition and a pure write:
\begin{equation}
 \begin{aligned}
 S_t&=A_tS_{t-1}+B_t,\\
 A_t&=\gamma_t(I-\beta_tk_tk_t^{\top}),\qquad
 B_t=\beta_tk_tv_t^{\top}.
 \end{aligned}
 \label{eq:ab}
\end{equation}
$A_t$ applies forgetting and key-selective correction to the historical state; $B_t$ has rank at most one. Given the current activations, $B_t$ contains no explicit dependence on this head's historical state, although those activations depend on the model and generated text during free generation.

\paragraph{Quantized writeback.}
\label{sec:state-quantization}
Let $\Q_b$ denote quantization followed by reconstruction at bit width $b$, and let $\hat S_{t-1}$ be the matrix decoded from the previous quantized state. Naive stepwise full-state quantization updates this matrix to the candidate state $S_t^{\mathrm{cand}}$, reads the current output, and then quantizes the state for the next step:
\begin{gather}
 S_t^{\mathrm{cand}}=A_t\hat S_{t-1}+B_t,
 \label{eq:state-candidate}\\
 o_t=(S_t^{\mathrm{cand}})^{\top}q_t,\qquad
 \hat S_t=\Q_b(S_t^{\mathrm{cand}}).
 \label{eq:writeback}
\end{gather}
$S_t^{\mathrm{cand}}$ already contains historical quantization errors. The new writeback error is introduced after the current readout and affects subsequent states and predictions.

\subsection{Recurrent-State Quantization}

\paragraph{Precision allocation.}
DAMP combines local quantization error and future decay to rank key channels, retaining a fixed number of high-risk channels per head in FP16 and quantizing the rest to INT8~\cite{damp}. Channels with similar error magnitudes can receive different protection because they are forgotten at different rates. For delta-rule states, STEPQuant uses heads as allocation units and combines error-persistence weights with candidate reconstruction distortion under a model-wide bit budget~\cite{stepquant}. Its distortion includes key-row readout weights estimated from propagated queries, while dual-axis scale fitting adapts to both axes of the state matrix. These approaches capture the magnitude, persistence, and local readout effects of state errors. Loss-based sensitivity has also been used for weight quantization: SqueezeLLM weights reconstruction errors with a diagonal Fisher approximation from gradient statistics and performs sensitivity-weighted nonuniform clustering~\cite{squeezellm}. Its targets are model weights that remain fixed after offline processing. Recurrent states instead change with each input, and each writeback perturbation undergoes further state transitions and readouts. GGA uses gradients of future language-model loss to weight candidate state residuals along this recurrence, retaining STEPQuant's candidate quantizers and nominal budget.

\paragraph{Delayed quantization.}
LeapQuant maintains a low-bit boundary state and a high-precision update buffer within each block window, recovers intermediate readouts using state-dependent net-correction terms, and reconstructs and requantizes the state at the boundary~\cite{leapquant}. Its Compensator Tokens retain dominant state structure as high-precision low-rank components across windows while quantizing the residual. Unlike buffering independent KV entries, a recurrent update buffer must account for transformations of earlier write contributions. PW separates the pure writes $B_t$ in Equation~\eqref{eq:ab}. Each recent write is stored as an FP8 vector pair whose left factor undergoes the full state transition; a rolling queue merges branches individually into the low-bit historical state according to write age. The old state remains updated and quantized at every step.

\section{Motivation}
\label{sec:motivation}
\suppressfloats[t]

\subsection{Matching Head Precision to Prediction Sensitivity}
\label{sec:allocation-observation}

\begin{figure}[!htbp]
\centering
\includegraphics[width=80mm]{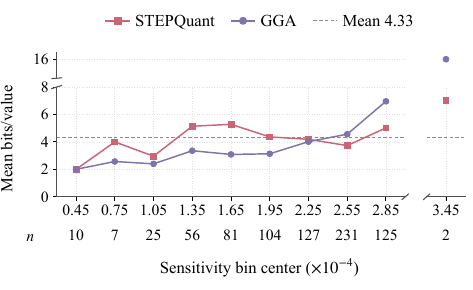}
\caption{Head precision versus empirical quantization impact. The horizontal axis shows centers of $0.30\times10^{-4}$-wide bins of $I_u$, with head counts $n$. Both allocations average 4.33 bit/value over all heads; FP16 counts as 16 bits. The empty bin $[3.00,3.30)\times10^{-4}$ and the vertical range 8--14 bits are omitted.}
\label{fig:allocation-impact}
\end{figure}

Assigning different precisions to recurrent-state heads should direct limited storage toward heads whose perturbations have a greater effect on subsequent predictions. We examine this correspondence on fixed diagnostic samples from Qwen3.5-4B. For each head, we introduce one quantization perturbation to its state, hold subsequent token inputs fixed, and measure the KL divergence between subsequent output distributions and the unquantized reference. Averaging this effect over the candidate precisions $\{2,4,6,8\}$ gives an empirical sensitivity $I_u$. Figure~\ref{fig:allocation-impact} groups heads into equal-width bins of $I_u$ and compares the mean bit width assigned to each bin; the horizontal axis runs from lower to higher empirical impact. Measurement details and complete binned data are provided in Appendix~\ref{app:allocation-diagnostic}.

The STEPQuant allocation does not increase consistently with this measured sensitivity. It assigns $5.28$ bits on average in the $[1.50,1.80)\times10^{-4}$ bin, but only $3.73$ bits in the higher-impact $[2.40,2.70)\times10^{-4}$ bin. STEPQuant combines error persistence, candidate reconstruction distortion, and key-row readout weights~\cite{stepquant}; these describe local properties of state perturbations without directly measuring the response of the final language-model loss. This observation motivates GGA: we introduce downstream loss gradients, propagate writeback sensitivity along reference state transitions, and use squared gradients to weight squared candidate quantization residuals before recomputing the model-wide allocation.

\FloatBarrier

\subsection{Preserving Recent Conclusions during Reasoning}
\suppressfloats[t]
\label{sec:pw-observation}

\begin{figure}[!htbp]
\centering
\begin{minipage}{\linewidth}
\fontsize{8}{9.5}\selectfont
\textbf{HMMT February 2026}\hfill\textcolor{obrgray}{Qwen3.5-9B; same prompt and seed}\par
\vspace{2pt}
Compute the sum of all positive integers $n$ such that $n$ has at least 6 positive integer divisors and the 6th largest divisor of $n$ is 6. \textbf{Reference answer: 48.}\par
\vspace{5pt}
\end{minipage}
\begin{minipage}[t][204pt][t]{.485\linewidth}
\fontsize{8}{9.5}\selectfont
\textcolor{obrred}{\textbf{GGA@$4$ (without PW)}}\par\vspace{2pt}\hrule height .3pt\vspace{4pt}
\begin{tabular}{@{}r@{\hspace{5pt}}p{.82\linewidth}@{}}
 & \textcolor{obrgray}{$\ldots$}\\[-2pt]
\textcolor{obrgray}{934} & Count: 1, 2, 3, 4, 5, 6, 7, 8, 9, 10.\\
\textcolor{obrgray}{935} & Okay, I miscounted when I said ``that is 9 numbers''.\\[-1pt]
 & \textcolor{obrgray}{$\ldots$}\\[-2pt]
\textcolor{obrgray}{947} & \textbf{Okay, there are 10 divisors.}\\
\textcolor{obrgray}{948} & \textcolor{obrred}{The list is 1, 2, 3, 4, 6, 8, 12, 16, 24, 48 is wrong.}\\
\textcolor{obrgray}{949} & \textcolor{obrred}{\textbf{Missing divisor:}}\\[-1pt]
 & \textcolor{obrgray}{$\ldots$}\\[-2pt]
\textcolor{obrgray}{962} & \textcolor{obrred}{\textbf{Wait, where is the 10th one?}}\\[-1pt]
 & \textcolor{obrgray}{$\ldots$}\\[-2pt]
\textcolor{obrgray}{1933} & Is there a factor 2? Yes.\\[-1pt]
 & \textcolor{obrgray}{$\ldots$}\\[-2pt]
\textcolor{obrgray}{1940} & \textcolor{obrred}{\textbf{Wait, did I forget $2^1\cdot3^0=2$? Yes.}}\\[-1pt]
 & \textcolor{obrgray}{$\ldots$}\\[-2pt]
\textcolor{obrgray}{5899} & \textcolor{obrred}{\textbf{Wait.}}\\
\textcolor{obrgray}{5900} & \textcolor{obrred}{\textbf{Wait.}}\\[-1pt]
 & \textcolor{obrgray}{$\ldots$}\\[-2pt]
\textcolor{obrgray}{6729} & \textcolor{obrred}{\textbf{Wait.}}\\
\end{tabular}
\vfill\vspace{3pt}\hrule height .3pt\vspace{3pt}
\textbf{81,921 tokens; repetition stop}\par
\textcolor{obrgray}{831 consecutive ``Wait.'' lines}
\end{minipage}\hfill
\begin{minipage}[t][204pt][t]{.485\linewidth}
\fontsize{8}{9.5}\selectfont
\textcolor{obrpurple}{\textbf{GGA@$4$ + PW-S8}}\par\vspace{2pt}\hrule height .3pt\vspace{4pt}
\begin{tabular}{@{}r@{\hspace{5pt}}p{.82\linewidth}@{}}
 & \textcolor{obrgray}{$\ldots$}\\[-2pt]
\textcolor{obrgray}{758} & Divisors of 48: 1, 2, 3, 4, 6, 8, 12, 16, 24, 48.\\[-1pt]
 & \textcolor{obrgray}{$\ldots$}\\[-2pt]
\textcolor{obrgray}{771} & \textcolor{obrpurple}{\textbf{Total $k=10$ divisors.}}\\[-1pt]
 & \textcolor{obrgray}{$\ldots$}\\[-2pt]
\textcolor{obrgray}{779} & So for $n=48$, the 6th largest divisor is indeed 6.\\
\textcolor{obrgray}{780} & So $n=48$ is a solution.\\[-1pt]
 & \textcolor{obrgray}{$\ldots$}\\[-2pt]
\textcolor{obrgray}{912} & \textcolor{obrpurple}{\textbf{Plus 6 itself. Total 10. Correct.}}\\
\textcolor{obrgray}{913} & So condition is satisfied.\\[-1pt]
 & \textcolor{obrgray}{$\ldots$}\\[-2pt]
\textcolor{obrgray}{915} & Now, let's explore general $n$.\\[-1pt]
 & \textcolor{obrgray}{$\ldots$}\\[-2pt]
\textcolor{obrgray}{2641} & And we concluded $n=48$ is the only solution.\\[-1pt]
 & \textcolor{obrgray}{$\ldots$}\\[-2pt]
\textcolor{obrgray}{2643} & \textcolor{obrpurple}{\textbf{Therefore the sum is 48.}}\\
\textcolor{obrgray}{2644} & \textcolor{obrgray}{</think>}\\[-1pt]
 & \textcolor{obrgray}{$\ldots$}\\[-2pt]
\textcolor{obrgray}{2677} & $\boxed{48}$\\
\end{tabular}
\vfill\vspace{3pt}\hrule height .3pt\vspace{3pt}
\textbf{57,682 tokens; natural EOS}\par
\textcolor{obrgray}{Final answer: 48}
\end{minipage}
\caption{Paired reasoning excerpts from the same problem. Numbers identify original output lines; ellipses mark omitted intervening lines. Without PW, the model rejects a verified divisor count, rechecks a known factor, and eventually repeats ``Wait.'' With PW, it retains the count, continues the search, and returns 48.}
\label{fig:pw-recent-conclusion}
\end{figure}

The left side of Figure~\ref{fig:pw-recent-conclusion} illustrates a failure to carry recently established conclusions forward under GGA@$4$. The problem asks for the sum of all positive integers with at least six positive divisors whose sixth-largest positive divisor is 6; the reference answer is 48. The model correctly lists all ten divisors of 48 and confirms their count on line 947, yet rejects the same complete list on line 948 and resumes searching for divisors that are not missing. The count just verified is no longer used consistently as the reasoning proceeds. Under stepwise full-state quantization, the state contribution supporting such a conclusion is incorporated into low-bit writeback immediately after the current readout, then repeatedly quantized with the accumulated history. Perturbing that recent contribution may affect the model's subsequent use of the conclusion.

The same trajectory acknowledges 2 as a divisor on line 1933 but again treats it as missing on line 1940. Rechecking eventually gives way to 831 consecutive lines of ``Wait.'' Generation is stopped by the repetition detector after $81{,}921$ tokens, without a final answer. These observations motivate preserving recent writes separately from the repeatedly quantized historical state. Writeback perturbations enter later readouts through the recurrence, and the text they influence becomes subsequent model input; this feedback offers a possible explanation for how stalled reasoning can persist. We therefore design PW to maintain recent pure writes separately and delay their incorporation into the low-bit accumulated state. Its representation and rolling update are developed in Section~\ref{sec:pw}, followed by the corresponding reasoning on the right of Figure~\ref{fig:pw-recent-conclusion}.

\section{Method}
\label{sec:method}
\suppressfloats[t]

\subsection{Overview}
\label{sec:overview}
We address recurrent-state quantization through precision allocation across heads and maintenance of recent writes over time. GGA constructs a loss-sensitive cost for each head and candidate bit width, then optimizes a model-wide mixed-precision allocation. PW separates recent pure writes from the accumulated history and delays their incorporation into the low-bit old state using a fixed-length queue.

GGA determines head bit widths and FP16 pivots offline; PW maintains recent branches online as FP8 vector pairs and applies the full GDN state transition to them. Their combination leaves model weights unchanged and requires no backpropagation during deployment. We present GGA in Section~\ref{sec:gga} and PW in Section~\ref{sec:pw}.

\subsection{Global Gradient-aware Precision Allocation}
\label{sec:gga}

\subsubsection{Sensitivity of Future Loss to the Writeback State}

To address the precision-allocation issue in Section~\ref{sec:allocation-observation}, GGA uses gradients of the language-model loss to construct candidate quantization risks and allocates bit widths across the entire model. The candidate quantizers, bit-width menu, and nominal budget follow STEPQuant.

Let $u\in\U$ denote a layer/head allocation unit. On an unquantized calibration trajectory, we use the next-token language-model loss
\begin{equation}
 \mathcal L=\sum_{t=1}^{T-1}
 -\log p_{\theta}(x_{t+1}\mid x_{\le t}).
 \label{eq:loss}
\end{equation}
Let $g_{u,t}=\partial\mathcal L/\partial o_{u,t}\in\mathbb R^{d_v}$ denote the downstream gradient of the loss with respect to the raw head readout. Since $o_{u,t}=S_{u,t}^{\top}q_{u,t}$, the current readout branch contributes $q_{u,t}g_{u,t}^{\top}$ to the state gradient.

To match the writeback position in Equation~\eqref{eq:writeback}, we distinguish the current readout contribution from the future recurrence contribution. Let $G_{u,t}^{\mathrm{wb}}$ denote the writeback-sensitivity adjoint and $J_{u,t}$ the state adjoint including the current readout. We propagate them backward along the recorded reference state transitions:
\begin{equation}
 \begin{aligned}
 J_{u,t}&=q_{u,t}g_{u,t}^{\top}+G_{u,t}^{\mathrm{wb}},\\
 G_{u,t-1}^{\mathrm{wb}}&=A_{u,t}^{\top}J_{u,t}.
 \end{aligned}
 \label{eq:adjoint}
\end{equation}
This backward recurrence is performed independently for each calibration segment. At the end of a segment, if the writeback state is no longer used by subsequent computation, then $G_{u,T}^{\mathrm{wb}}=0$. This boundary condition is consistent with the actual range of supervised tokens.

Equation~\eqref{eq:adjoint} describes adjoint propagation on the explicit state-recurrence graph, given the reference activations of this head; the downstream gradient $g_{u,t}$ supplies sensitivity information from the language-model loss. We use $G_{u,t}^{\mathrm{wb}}$ as a sensitivity estimate for candidate quantization risk. Adjoint propagation includes the forgetting term and key-selective correction in $A_{u,t}$. Because the current writeback occurs after the current readout, we use $G_{u,t}^{\mathrm{wb}}$ to construct the cost, without attributing the already completed readout contribution to the newly introduced error.

\subsubsection{Squared-Gradient-Weighted Candidate Cost}

For a high-precision snapshot $S_{u,t}^{\mathrm{FP}}$ of head $u$, the candidate quantization residual is
\begin{equation}
 \Delta S_{u,t}^{(b)}
 =\Q_b^{\mathrm{STEP}}(S_{u,t}^{\mathrm{FP}})-S_{u,t}^{\mathrm{FP}}.
 \label{eq:residual}
\end{equation}
Let $\mathcal T_u$ be the set of valid snapshots across calibration segments, jointly indexed by segment and time, with the segment index omitted below; $N_u=|\mathcal T_u|>0$ is the total number of snapshots. GGA defines the nonnegative cost
\begin{equation}
 C_{u,b}^{\mathrm{GGA}}
 =\frac{1}{2N_u}
 \sum_{t\in\mathcal T_u}
 \left\|G_{u,t}^{\mathrm{wb}}\odot\Delta S_{u,t}^{(b)}\right\|_F^2.
 \label{eq:gga-cost}
\end{equation}
Here, $\odot$ denotes elementwise multiplication, and $\|\cdot\|_F$ is the Frobenius norm. This cost assigns greater weight to quantization residuals at loss-sensitive coordinates and averages over the calibration snapshots. By comparison, the STEP allocation cost combines a lifetime weight with candidate reconstruction distortion:
\begin{equation}
 C_{u,b}^{\mathrm{STEP}}=L_u\,d_u(b).
 \label{eq:step-cost}
\end{equation}
$d_u(b)$ follows the original STEP definitions of candidate quantization and distortion. GGA does not additionally multiply Equation~\eqref{eq:gga-cost} by $L_u$, nor does it normalize costs separately for different layers or heads, preserving their global relative scale.

Equation~\eqref{eq:gga-cost} uses diagonal sensitivity weighting on the reference trajectory to aggregate the single-perturbation quantization residuals at individual coordinates into an approximate head-level risk. The sum of these head costs then forms the global allocation objective.

\subsubsection{Global Allocation under a Nominal Budget}

Let $\mathcal B$ be the set of candidate integer bit widths and $\bar b$ the nominal average bit-width budget. GGA follows STEPQuant's allocation framework, retaining a small number of heads in FP16, called pivots, and selecting an integer bit width from $\mathcal B$ for each remaining head. For state heads of identical shape, we first solve a preliminary allocation over all heads without pivots, under the budget $\bar b|\U|$, obtaining bit widths $b_u^{(0)}$ and their corresponding residual risks $r_u$:
\begin{equation}
 \begin{alignedat}{2}
 \{b_u^{(0)}\}&\in\arg\min_{\{b_u\}}
       \sum_{u\in\U}C_{u,b_u}^{\mathrm{GGA}},\qquad &
 r_u&=C_{u,b_u^{(0)}}^{\mathrm{GGA}},\\
 \mathrm{s.t.}\quad &b_u\in\mathcal B,\quad
       \sum_{u\in\U}b_u=\bar b|\U|.
 \end{alignedat}
 \label{eq:pivot-risk}
\end{equation}
We select a prescribed number $P$ of heads as pivots in descending order of their residual risks $r_u$ after the preliminary allocation, and denote their set by $\Piv$. After fixing this set, let $B_{\mathrm{int}}$ be the budget for the sum of integer bit widths available to the remaining non-pivot heads. We reallocate these heads as follows:
\begin{equation}
 \begin{aligned}
 \min_{\{b_u\}}\quad &
 \sum_{u\notin\Piv}C_{u,b_u}^{\mathrm{GGA}}\\
 \mathrm{s.t.}\quad &b_u\in\mathcal B,\quad
 \sum_{u\notin\Piv}b_u=B_{\mathrm{int}}.
 \end{aligned}
 \label{eq:allocation}
\end{equation}
Both stages are solved by global dynamic programming; the recurrence and backtracking procedure are given in Appendix~\ref{app:allocation-dp}. Bit widths are allocated across the entire model, with each layer's share determined jointly by the candidate costs and the total budget.

STEP and GGA use the same nominal constraint, while the specific pivot identities and integer bit-width allocations may differ. Heterogeneous state shapes require corresponding budgets based on the number of elements. Gradients and candidate costs are used only for offline calibration; deployment directly uses the frozen pivot set and bit-width table, without backpropagation. Candidate bit widths and pivot counts are given in Section~\ref{sec:experimental-setup}, and calibration and frozen configurations in Appendix~\ref{app:implementation}.

\paragraph{Resulting precision distribution.}
Figure~\ref{fig:allocation-impact} shows how the GGA allocation responds to the mismatch identified in Section~\ref{sec:allocation-observation}. At the same all-head average of $4.33$ bit/value, GGA's mean bit width rises overall with empirical quantization impact. It assigns $4.56$ versus STEPQuant's $3.73$ bits in the $[2.40,2.70)\times10^{-4}$ bin, and $6.96$ versus $5.02$ bits in the $[2.70,3.00)\times10^{-4}$ bin. Lower-impact bins receive fewer bits on average, directing more precision toward head groups with greater measured effects on subsequent predictions.

\subsection{Pure-Write Sliding Window}
\label{sec:pw}

\subsubsection{Exact Expansion and Vector Representation of Recent Writes}

To address the maintenance of recent information discussed in Section~\ref{sec:pw-observation}, PW separates the pure-write branches in Equation~\eqref{eq:ab}, maintains the most recent $m$ writes as FP8 vector pairs, and propagates them through the full state transition. Earlier history is carried by the stepwise-quantized old state.

\begin{figure}[!htbp]
\centering
\includegraphics[width=128mm]{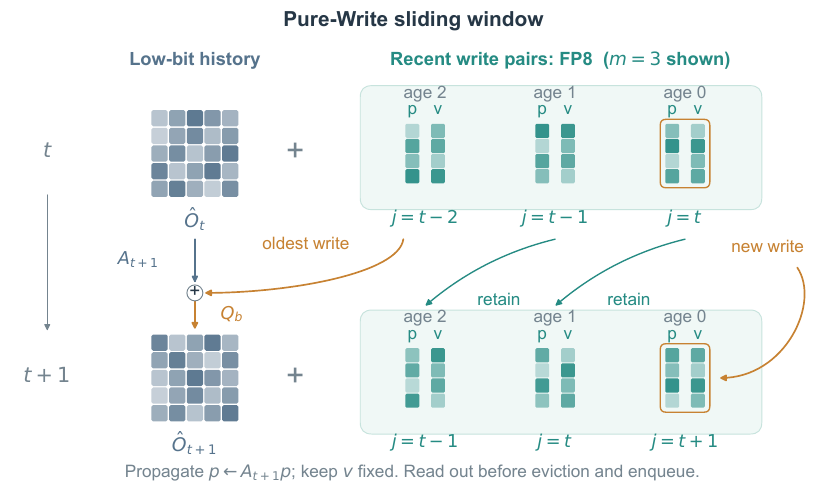}
\caption{PW state decomposition and rolling update, with $m=3$ shown for illustration. Each FP8 vector pair contributes a rank-one write, maintained separately from the low-bit history. Existing left factors and the history propagate through $A_{t+1}$. After readout, the oldest branch is merged into history before requantization and the new write enters the window; retained right factors remain fixed.}
\label{fig:pw-concept}
\end{figure}

Figure~\ref{fig:pw-concept} illustrates the state decomposition and the one-step shift of the recent-write window.

Let $r=\min(m,t)$ and unroll Equation~\eqref{eq:ab} over the most recent $r$ steps:
\begin{equation}
 S_t=(A_t\cdots A_{t-r+1})S_{t-r}
 +\sum_{j=t-r+1}^{t}(A_t\cdots A_{j+1})B_j.
 \label{eq:unroll}
\end{equation}
An empty product is defined as the identity matrix. The first term is the earlier state propagated to the current time, while the second term is the propagated contribution of recent pure writes.

Since $B_j=(\beta_jk_j)v_j^{\top}$, we define a left factor for each branch:
\begin{equation}
 p_{j,j}=\beta_jk_j,\qquad p_{j,t}=A_tp_{j,t-1}.
 \label{eq:p-def}
\end{equation}
Then
\begin{equation}
 (A_t\cdots A_{j+1})B_j=p_{j,t}v_j^{\top},\qquad
 p_{j,t}=\gamma_t\bigl[p_{j,t-1}
       -\beta_tk_t(k_t^{\top}p_{j,t-1})\bigr].
 \label{eq:p-update}
\end{equation}
Left multiplication by $A_t$ preserves the rank-one structure, so each recent branch only needs to store $(p_{j,t},v_j)$ rather than a full state matrix. $v_j$ is the value stored at the birth of the branch and is not updated by subsequent transitions; their effects accumulate in $p_{j,t}$.

\subsubsection{Online State Maintenance and Readout Timing}

Let $\hat O_t$ be the reconstructed matrix of the low-bit old state. After the update, the queue index set is
\begin{equation}
 \mathcal J_t=\{\max(1,t-m+1),\ldots,t\},\qquad
 \hat S_t=\hat O_t+\sum_{j\in\mathcal J_t}p_{j,t}v_j^{\top}.
 \label{eq:split}
\end{equation}
The sum here defines the logical state; the implementation does not persistently materialize the entire recent matrix. The queue is initially empty at the decoding boundary, i.e., $\mathcal J_0=\varnothing$, and the old state is initialized from the persistent-state representation produced by the original prefill path.

At each decoding update, we first propagate the old state and all existing recent left factors:
\begin{equation}
 O_t^{\mathrm{pre}}=A_t\hat O_{t-1},\qquad
 p_{j,t}=A_tp_{j,t-1},\quad j\in\mathcal J_{t-1}.
 \label{eq:pw-propagate}
\end{equation}
The current new write is $p_{t,t}=\beta_tk_t$. The current readout uses the old state, all propagated recent branches, and the current new write:
\begin{equation}
 o_t=(O_t^{\mathrm{pre}})^{\top}q_t
 +\sum_{j\in\mathcal J_{t-1}}v_j(p_{j,t}^{\top}q_t)
 +v_t\bigl((\beta_tk_t)^{\top}q_t\bigr).
 \label{eq:pw-read}
\end{equation}
If the queue is full upon entering this step, the oldest branch $j=t-m$ is evicted after the current readout and merged into the old state. Define
\begin{equation}
 E_t^{\mathrm{evict}}=
 \begin{cases}
 p_{t-m,t}v_{t-m}^{\top},&t>m,\\
 0,&t\le m.
 \end{cases}
 \label{eq:evict}
\end{equation}
The merged old state is then quantized, and the current new write is added to the queue:
\begin{equation}
 O_t^{\mathrm{merge}}=O_t^{\mathrm{pre}}+E_t^{\mathrm{evict}},\qquad
 \hat O_t=\Q_{b_u}(O_t^{\mathrm{merge}}).
 \label{eq:pw-write}
\end{equation}
When the queue is full, the current readout has access to the original $m$ historical branches and the current write; the persistent queue after the update still contains at most $m$ vector pairs. The old state is updated and quantized at every step. PW delays the first incorporation of a new write into the low-bit old state. Before the window fills, there is no eviction term, but the old state still undergoes its prescribed update and writeback.

The recent component has rank at most $m$ and is stored separately as FP8 vector pairs. Recurrence equivalence under ideal precision is established in Appendix~\ref{app:equivalence}, and the FP8 encoding and complete update algorithm are given in Appendix~\ref{app:pw-implementation}; combined configurations and equivalent state bit widths are described in Appendix~\ref{app:storage}.

\paragraph{Case-study response.}
The right side of Figure~\ref{fig:pw-recent-conclusion} presents the generation with PW-S8, using the same prompt, random seed, sampling parameters, and GGA@$4$ bit-width configuration as the trajectory in Section~\ref{sec:pw-observation}. After rechecking, it carries forward that there are ten divisors and the sixth-largest is six, examines other candidates, and terminates naturally with the answer 48. The complete output contains $57{,}682$ tokens, $24{,}239$ fewer than without PW, a reduction of $29.59\%$. Section~\ref{sec:experiments} evaluates the effects on scores and generation length across tasks.

\FloatBarrier

\section{Experiments}
\label{sec:experiments}
\suppressfloats[t]
This section reports task performance, component comparisons, and generation behavior, and compares persistent-state storage costs in equivalent bits.

\subsection{Experimental Setup}
\label{sec:experimental-setup}

\paragraph{Models and comparison methods.}
We evaluate post-trained Qwen3.5-4B and Qwen3.5-9B~\cite{qwen35-4b,qwen35-9b}. Each evaluated model contains 24 GDN layers, with 32 state/value heads of size $128\times128$ per layer and 768 heads in total; weights and convolutions remain in BF16. The @$4$ group compares naive INT4, STEPQuant@$4$, GGA@$4$, and each allocation combined with PW-S8, alongside FP32 and FP16 recurrent-state references. The @$6$ group compares naive INT6, STEPQuant@$6$, GGA@$6$, and each allocation combined with PW-S8; we additionally compare STEP/GGA@$5$ against @$4$+PW-S8 to examine how similar storage budgets are used. Naive integer states use symmetric quantization with a group size of 128 along the value axis; STEP and GGA use the same candidate quantizers, with a value group size of 32. The candidate integer bit widths for both @$4$ and @$5$ are $\{2,4,6,8\}$ bits, and those for @$6$ are $\{4,6,8\}$ bits; all budgets retain 32 FP16 heads. PW-S8 denotes a window length of $m=8$, with the queue format defined in Section~\ref{sec:pw}; each combination uses the frozen precision configuration at its corresponding nominal budget. Floating-point references accumulate in FP32 and persistently store the recurrent state in the target dtype. State storage is measured in equivalent bits/value, including state codes, scales, FP16 pivots, and the PW queue; the accounting method and complete storage-cost table are provided in Appendix~\ref{app:storage}.

\paragraph{Calibration.}
We perform offline calibration on 32 segments from the WikiText-2 training set, each containing 2048 tokens, to estimate candidate quantization errors, readout effects, and loss sensitivity, and to determine head bit widths and FP16 pivots. Calibration details are provided in Appendix~\ref{app:implementation}.

\paragraph{Evaluation data.}
We evaluate on randomly sampled evaluation sets drawn from AIME 2026 (30 problems), HMMT February 2026 (33 problems)~\cite{matharena}, GPQA Diamond (198 questions)~\cite{gpqa}, HumanEval+ (164 problems)~\cite{evalplus}, MATH500 (500 problems)~\cite{math500}, IFBench (300 prompts)~\cite{ifbench}, and LiveCodeBench v6 (200 problems)~\cite{livecodebench}. The problem lists are fixed before evaluation and shared across all comparison methods; we independently sample five responses per problem for AIME and HMMT, and two responses per problem for the remaining datasets.

\paragraph{Generation settings.}
All settings use the same frozen problem lists, prompts, and per-problem seeds, with temperature $=1$, top-$p=0.95$, top-$k=20$, thinking enabled, an output limit of 100,000 tokens (including thinking), and a context limit of 131,072. FP32/FP16 serve as floating-point state references; gains between methods are calculated only among completed quantized settings. Responses end naturally upon EOS; a repetition detector using the same rules also stops clearly periodic outputs, which are counted separately from outputs that reach the token limit. The rules are given in Appendix~\ref{app:evaluation-protocol}. Responses without a final answer remain in the evaluation denominator.

Inference uses an A100-SXM4-80GB GPU with SGLang 0.5.12.

\FloatBarrier

\subsection{Main Results}
\suppressfloats[t]
\label{sec:main-results}

\begin{figure}[!htbp]
\centering
\begin{minipage}[t]{.48\linewidth}
\centering
\includegraphics[width=66mm]{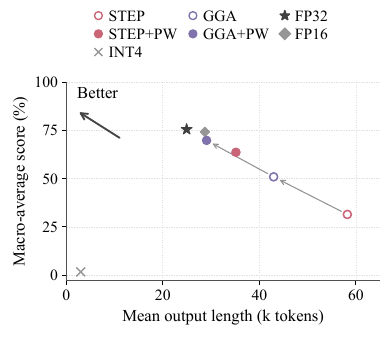}
\caption{Score--length tradeoff on Qwen3.5-4B at @$4$. Both axes weight tasks equally. Gray arrows trace STEP to GGA to GGA+PW; FP32 and FP16 show floating-point references. PW denotes PW-S8 and INT4 denotes naive INT4.}
\label{fig:accuracy-length-4b-4}
\end{minipage}
\hfill
\begin{minipage}[t]{.48\linewidth}
\centering
\includegraphics[width=66mm]{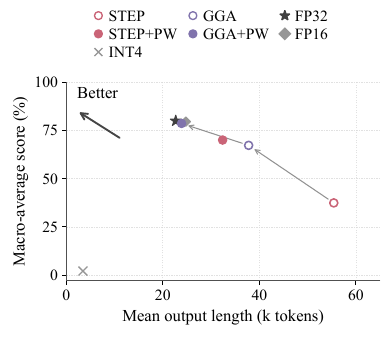}
\caption{Score--length tradeoff on Qwen3.5-9B at @$4$, with FP32/FP16 recurrent-state references and naive INT4. Markers, method arrows, and the better direction follow Figure~\ref{fig:accuracy-length-4b-4}.}
\label{fig:accuracy-length-9b-4}
\end{minipage}
\end{figure}

Table~\ref{tab:main-results} compares task scores for floating-point state references and the @$4$ and @$6$ settings. Figures~\ref{fig:accuracy-length-4b-4} and~\ref{fig:accuracy-length-9b-4} summarize the @$4$ score--length tradeoff; task-level lengths are provided in Appendix~\ref{app:evaluation-protocol}, Table~\ref{tab:task-lengths}. At both budgets, the benefits of GGA and PW appear in task performance and generation behavior: loss-aware allocation improves the use of limited bit widths, and recent-write maintenance further improves scores and mitigates excessively long outputs.

\begin{table}[!htbp]
\centering
\fontsize{8.5}{10}\selectfont
\setlength{\tabcolsep}{3pt}
\caption{Measured scores (\%) on seven tasks at @$4$ and @$6$. For each model, floating-point references are followed by the @$4$ and @$6$ groups; boldface compares scores among completed quantized settings within each model and nominal-budget group. Scores are averaged across independent samples, and the macro-average weights all tasks equally. HE+, IFB, and LCB denote HumanEval+ base+plus, IFBench prompt strict, and the LiveCodeBench v6 evaluation subset, respectively.}
\label{tab:main-results}
\begin{tabular*}{\textwidth}{@{\extracolsep{\fill}}llrrrrrrrr@{}}
\toprule
Model & Method & AIME & HMMT & GPQA & HE+ & MATH & IFB & LCB & Macro-avg.\\
\midrule
4B & FP32 & 86.67 & 63.64 & 70.96 & 86.89 & 94.90 & 62.00 & 63.75 & 75.54\\
 & FP16 & 84.00 & 61.82 & 71.97 & 86.28 & 95.60 & 58.00 & 62.25 & 74.27\\
\cmidrule(lr){2-10}
 & Naive INT4 & 0.00 & 0.00 & 1.26 & 0.00 & 2.40 & 8.00 & 0.00 & 1.67\\
 & STEP@$4$ & 0.67 & 0.00 & 16.41 & 78.96 & 47.30 & 46.50 & 30.25 & 31.44\\
 & GGA@$4$ & 30.67 & 17.58 & 50.51 & 78.66 & 83.60 & 51.67 & 43.75 & 50.92\\
 & STEP@$4$+PW-S8 & 54.67 & 37.58 & 60.61 & \textbf{82.62} & 94.00 & 56.00 & \textbf{60.25} & 63.67\\
 & GGA@$4$+PW-S8 & \textbf{74.00} & \textbf{57.58} & \textbf{62.12} & 81.10 & \textbf{97.30} & \textbf{56.50} & \textbf{60.25} & \textbf{69.83}\\
\cmidrule(lr){2-10}
 & Naive INT6 & 0.00 & 0.00 & 1.52 & 13.11 & 13.80 & 17.83 & 2.25 & 6.93\\
 & STEP@$6$ & 74.67 & 51.52 & 65.66 & 83.54 & 96.00 & \textbf{61.17} & 61.25 & 70.54\\
 & GGA@$6$ & 78.00 & 57.58 & 64.65 & \textbf{85.37} & 97.30 & 59.17 & 60.75 & 71.83\\
 & STEP@$6$+PW-S8 & \textbf{83.33} & 63.03 & 67.17 & 83.54 & \textbf{98.20} & 59.00 & \textbf{61.50} & 73.68\\
 & GGA@$6$+PW-S8 & 82.00 & \textbf{65.45} & \textbf{69.70} & 82.93 & 97.90 & 58.50 & 61.25 & \textbf{73.96}\\
\midrule
9B & FP32 & 90.67 & 66.67 & 76.77 & 87.20 & 97.50 & 64.83 & 76.00 & 79.95\\
 & FP16 & 88.67 & 70.30 & 76.01 & 86.89 & 96.70 & 64.50 & 74.00 & 79.58\\
\cmidrule(lr){2-10}
 & Naive INT4 & 0.00 & 0.00 & 1.52 & 0.61 & 1.60 & 11.00 & 0.00 & 2.10\\
 & STEP@$4$ & 2.67 & 4.24 & 25.25 & 83.84 & 59.10 & 48.83 & 38.00 & 37.42\\
 & GGA@$4$ & 66.00 & 45.45 & 66.41 & 82.62 & 94.90 & 58.17 & 57.25 & 67.26\\
 & STEP@$4$+PW-S8 & 67.33 & 46.67 & 66.41 & \textbf{87.50} & 96.50 & 60.00 & 65.75 & 70.02\\
 & GGA@$4$+PW-S8 & \textbf{84.00} & \textbf{72.12} & \textbf{69.70} & 86.89 & \textbf{99.30} & \textbf{62.67} & \textbf{75.75} & \textbf{78.63}\\
\cmidrule(lr){2-10}
 & Naive INT6 & 0.00 & 0.00 & 5.05 & 16.77 & 24.70 & 23.50 & 6.00 & 10.86\\
 & STEP@$6$ & 80.00 & 63.64 & 70.20 & \textbf{88.11} & 97.90 & \textbf{66.33} & 72.75 & 76.99\\
 & GGA@$6$ & \textbf{86.67} & 66.06 & 75.00 & \textbf{88.11} & 98.80 & 65.17 & 74.75 & 79.22\\
 & STEP@$6$+PW-S8 & 86.00 & \textbf{67.27} & 74.24 & 85.06 & \textbf{99.20} & 64.50 & 74.50 & 78.68\\
 & GGA@$6$+PW-S8 & 84.00 & \textbf{67.27} & \textbf{79.80} & \textbf{88.11} & 98.80 & 65.67 & \textbf{75.50} & \textbf{79.88}\\
\bottomrule
\end{tabular*}
\end{table}

\paragraph{Effects of low-bit state quantization on long-reasoning tasks.}
At @$4$, naive INT4 achieves macro-averages of only $1.67\%$/$2.10\%$ on 4B/9B, indicating that directly applying uniform low-bit quantization to the full recurrent state severely degrades model capabilities. STEPQuant's mixed-precision scheme achieves macro-averages of $31.44\%$/$37.42\%$, but the extent of recovery varies substantially across tasks. For example, its HumanEval+ scores remain at $78.96\%$/$83.84\%$, whereas AIME scores are only $0.67\%$/$2.67\%$, and HMMT scores are only $0.00\%$/$4.24\%$. Long-reasoning tasks are particularly affected, and average scores alone do not fully reflect this degradation. AIME and HMMT therefore provide important tasks for examining precision allocation and recent-information maintenance.

\paragraph{GGA recovers task performance under the same budget.}
Keeping the candidate quantizers and nominal budget unchanged and replacing only STEP allocation with GGA improves the 4B/9B macro-averages by $19.48$/$29.84$ percentage points, to $50.92\%$/$67.26\%$. This improvement is not confined to a single problem set: on 9B, for example, GPQA increases from $25.25\%$ to $66.41\%$, MATH500 from $59.10\%$ to $94.90\%$, and AIME and HMMT recover to $66.00\%$ and $45.45\%$, respectively. These improvements under the same bit-width budget indicate that precision configuration across heads itself has a substantial effect on task performance. Together with the allocation observations in Section~\ref{sec:allocation-observation}, these results suggest that sensitivity to final prediction loss provides a more effective basis for bit-width allocation, directing limited precision toward states with greater effects on predictions.

\paragraph{PW further improves existing precision allocations.}
Adding PW-S8 to GGA@$4$ further raises the 4B/9B macro-averages to $69.83\%$/$78.63\%$, improvements of $18.92$/$11.37$ percentage points over GGA@$4$, respectively. On 9B, AIME increases from $66.00\%$ to $84.00\%$, HMMT from $45.45\%$ to $72.12\%$, and the macro-average approaches the FP16 state reference of $79.58\%$. PW is also effective with STEP allocation: STEP@$4$+PW-S8 achieves macro-averages of $63.67\%$/$70.02\%$. With PW-S8 fixed, GGA still improves over STEP allocation by $6.16$/$8.61$ percentage points. Together, these comparisons demonstrate the complementary roles of the two designs: GGA improves precision configuration across heads, while PW protects recent writes during state updates, enabling existing precision allocations to better support sustained reasoning. Section~\ref{sec:budget-comparison} compares the additional storage introduced by the window with directly increasing bit widths.

\paragraph{Allocation and maintenance remain effective at @$6$.}
With a larger state budget, STEPQuant@$6$ recovers 4B/9B macro-averages to $70.54\%$/$76.99\%$, improvements of $39.10$/$39.57$ percentage points over @$4$; AIME recovers to $74.67\%$/$80.00\%$ and HMMT to $51.52\%$/$63.64\%$. This indicates that increasing bit widths substantially mitigates the degradation of long reasoning at low budgets, and provides a comparison for examining the methods after capabilities have recovered. GGA@$6$ achieves macro-averages of $71.83\%$/$79.22\%$, still exceeding STEPQuant@$6$ by $1.29$/$2.23$ percentage points. The score gap between the two allocations narrows relative to @$4$, but the benefit of loss-aware allocation remains on both models.

\begin{figure}[!htbp]
\centering
\begin{minipage}[t]{.48\linewidth}
\centering
\includegraphics[width=66mm]{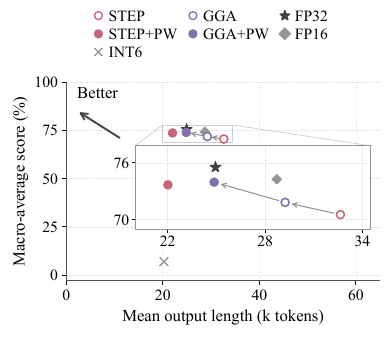}
\caption{Score--length tradeoff on Qwen3.5-4B at @$6$. The inset enlarges the high-score configurations, including the FP32/FP16 references and method arrows. PW denotes PW-S8 and INT6 denotes naive INT6.}
\label{fig:accuracy-length-4b-6}
\end{minipage}
\hfill
\begin{minipage}[t]{.48\linewidth}
\centering
\includegraphics[width=66mm]{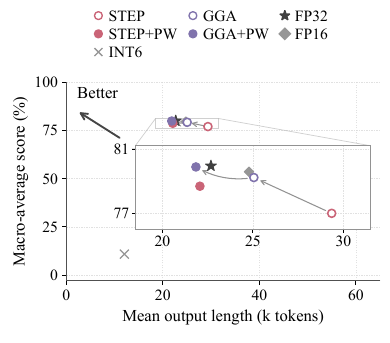}
\caption{Score--length tradeoff on Qwen3.5-9B at @$6$. The inset enlarges the high-score configurations; markers, method arrows, and the better direction use the same conventions as Figures~\ref{fig:accuracy-length-4b-4}--\ref{fig:accuracy-length-4b-6}.}
\label{fig:accuracy-length-9b-6}
\end{minipage}
\end{figure}

PW also remains effective at this higher budget. STEPQuant@$6$+PW-S8 reaches macro-averages of $73.68\%$/$78.68\%$, and GGA@$6$+PW-S8 reaches $73.96\%$/$79.88\%$. The GGA combination improves over GGA@$6$ by $2.13$/$0.66$ percentage points, approaching the overall performance of the floating-point state references. Thus, the value of recent-write maintenance extends beyond recovering severely degraded capabilities at @$4$ to further improving task performance at higher precision. Meanwhile, differences in generation length at similar scores can further reveal how different state-maintenance strategies affect reasoning. Figures~\ref{fig:accuracy-length-4b-6} and~\ref{fig:accuracy-length-9b-6} show the corresponding score--length comparisons, with close-ups of the high-score configurations.

\paragraph{GGA shortens mean outputs at both budgets.}
Table~\ref{tab:task-lengths} first computes the mean full output length for each task and then averages equally across the seven tasks, preventing tasks with more problems from dominating the overall result. At @$4$, GGA reduces the 4B/9B mean lengths from STEP's $58.23$/$55.42$ thousand tokens to $42.96$/$37.78$ thousand tokens, reductions of $26.23\%$/$31.83\%$. At @$6$, although STEPQuant has already substantially recovered task performance, GGA still reduces mean lengths from $32.64$/$29.35$ thousand tokens to $29.23$/$25.05$ thousand tokens, reductions of $10.45\%$/$14.65\%$. In both allocation comparisons, all seven tasks have shorter mean outputs on both models. The length improvements are therefore not driven by changes in only a few tasks; loss-aware allocation consistently mitigates excessive generation after quantization while improving task macro-average scores.

\paragraph{PW further improves output lengths in long reasoning.}
Adding PW-S8 to GGA@$4$ further reduces the 4B/9B mean lengths to $29.07$/$23.89$ thousand tokens, additional reductions of $32.32\%$/$36.77\%$ relative to GGA without a window. Adding PW-S8 to STEPQuant@$4$ also reduces mean lengths by $39.63\%$/$41.56\%$, respectively. Both allocation methods gain further benefits from the window, indicating that recent-write maintenance still has room for improvement after precision is reallocated. On 9B AIME, for example, STEPQuant@$4$, GGA@$4$, and GGA@$4$+PW-S8 produce mean lengths of $93.51$, $69.27$, and $43.23$ thousand tokens, respectively, while their scores increase from $2.67\%$ to $66.00\%$ and then to $84.00\%$. These changes show that higher accuracy accompanies shorter mean outputs, consistent with the case in Section~\ref{sec:pw}, where recent conclusions are retained and reasoning continues to advance.

PW's length benefits also extend to @$6$. STEPQuant@$6$+PW-S8 has 4B/9B mean lengths of $22.01$/$22.07$ thousand tokens, reductions of $32.58\%$/$24.80\%$ relative to the corresponding settings without a window; GGA@$6$+PW-S8 has mean lengths of $24.86$/$21.85$ thousand tokens, reductions of $14.98\%$/$12.76\%$ relative to GGA@$6$. Comparing the two window combinations, the GGA combination on 4B achieves a slightly higher macro-average score, whereas the STEP combination has shorter mean outputs; on 9B, the GGA combination achieves both a higher macro-average score and slightly shorter mean outputs. These results show that PW improves scores and mean lengths with both allocations, while the advantages of specific combinations vary with the model and budget.

\begin{figure}[!htbp]
\centering
\begin{minipage}[t]{.48\linewidth}
\centering
\includegraphics[width=66mm]{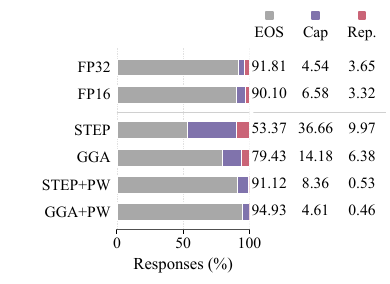}
\caption{Stopping categories on Qwen3.5-4B at @$4$. Natural EOS, the 100K cap (Cap), and repetition stops partition the responses; PW denotes PW-S8.}
\label{fig:termination-4b-4}
\end{minipage}
\hfill
\begin{minipage}[t]{.48\linewidth}
\centering
\includegraphics[width=66mm]{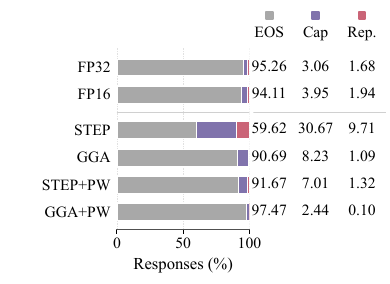}
\caption{Stopping categories on Qwen3.5-9B at @$4$, alongside floating-point state references.}
\label{fig:termination-9b-4}
\end{minipage}
\end{figure}

\paragraph{Shorter outputs accompany fewer abnormal stops.}
Figures~\ref{fig:termination-4b-4} and~\ref{fig:termination-9b-4} show the stopping categories at @$4$, further explaining the generation behavior behind the length improvements; complete results are provided in Appendix~\ref{app:evaluation-protocol}, Table~\ref{tab:termination}. At @$4$, $36.66\%$/$30.67\%$ of STEPQuant responses on 4B/9B reach the 100K limit, and another $9.97\%$/$9.71\%$ are stopped for repetitive output. With GGA@$4$+PW-S8, these rates decrease to $4.61\%$/$2.44\%$ and $0.46\%$/$0.10\%$, respectively, while natural EOS rates reach $94.93\%$/$97.47\%$. The reduction in mean output length thus accompanies fewer responses that exhaust the generation budget or fall into repetition, demonstrating effective mitigation of excessive generation.

Naive INT4 also produces very short outputs that all end with EOS, but its task scores are near zero; GGA and PW simultaneously improve task scores, shorten mean outputs, and reduce token-limit and repetition stops. This indicates that shorter outputs accompany the recovery of problem-solving capabilities and a higher proportion of completed responses. At @$6$, adding PW to either allocation also exhibits the joint trend of higher scores, shorter mean outputs, and fewer abnormal stops, further supporting the benefits of recent-write maintenance for quantized reasoning.

\FloatBarrier

\subsection{Supplementary Experiments: Higher Precision versus Window Maintenance at Similar Budgets}
\suppressfloats[t]
\label{sec:budget-comparison}

\begin{figure}[!htbp]
\centering
\begin{minipage}[t]{.48\linewidth}
\centering
\includegraphics[width=66mm]{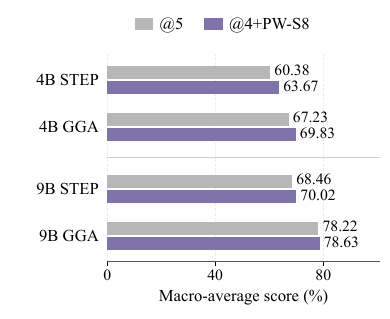}
\caption{Seven-task macro-average scores at similar storage budgets: @$5$ versus @$4$+PW-S8 under each allocation method. The two settings have similar equivalent state-storage costs (Table~\ref{tab:budget-comparison}).}
\label{fig:budget-score}
\end{minipage}
\hfill
\begin{minipage}[t]{.48\linewidth}
\centering
\includegraphics[width=66mm]{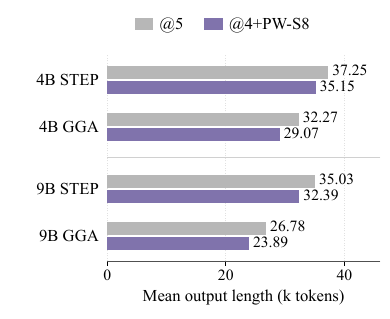}
\caption{Task-equal mean output lengths at similar storage budgets. The paired bars compare @$5$ with @$4$+PW-S8.}
\label{fig:budget-length}
\end{minipage}
\end{figure}

The main results show that both increasing bit widths and introducing PW improve task performance. A further question is how to use similar amounts of additional state storage: to increase the average bit width of the accumulated state, or to maintain recent writes independently? To examine this question, we pair @$5$ with @$4$+PW-S8 under the same allocation method: the former increases state bit widths, whereas the latter retains the @$4$ setting and adds a window. Figures~\ref{fig:budget-score} and~\ref{fig:budget-length} compare seven-task macro-averages and task-equal mean output lengths, respectively. The window configurations require $0.10$--$0.11$ more bits/value than the corresponding @$5$ configurations; complete numeric results are provided in Appendix~\ref{app:evaluation-protocol}, Table~\ref{tab:budget-comparison}.

\paragraph{Using the additional budget for window maintenance.}
Under STEP allocation, @$4$+PW-S8 improves 4B/9B macro-averages over @$5$ by $3.29$/$1.57$ percentage points and reduces mean output lengths by $5.63\%$/$7.54\%$, respectively. Under GGA allocation, the corresponding score improvements are $2.61$/$0.41$ percentage points, with mean-length reductions of $9.91\%$/$10.80\%$. Thus, in these comparisons at similar storage budgets, window maintenance achieves higher macro-average scores and shorter mean outputs with both allocations and both models, indicating that PW's benefits persist when compared with directly increasing bit widths.

On 9B, GGA@$5$ and GGA@$4$+PW-S8 achieve very similar macro-averages of $78.22\%$ and $78.63\%$, but the latter still reduces mean output length from $26.78$ thousand tokens to $23.89$ thousand tokens. This indicates that even when task scores approach the floating-point reference, recent-write maintenance provides additional benefits by mitigating excessive outputs. Appendix~\ref{app:evaluation-protocol} provides the per-task lengths and termination statistics for @$5$ in Tables~\ref{tab:budget-task-lengths} and~\ref{tab:supplementary-termination}, respectively.

Increasing state bit widths reduces quantization distortion in the accumulated state, whereas PW devotes additional storage to recent writes, delaying their entry into the low-bit historical state. These results support allocating part of the storage resources according to write age: beyond deciding how many bits each head should use, the time at which information is first subjected to low-bit quantization after being written also affects task performance and generation length.

\section{Discussion and Limitations}
\label{sec:limitations}
\suppressfloats[t]

GGA and PW improve recurrent-state quantization through precision allocation and write maintenance, respectively. GGA uses prediction-loss sensitivity to guide bit-width allocation across heads, improving task scores and shortening mean outputs on the evaluated models. PW independently maintains recent writes and delays their incorporation into the low-bit historical state, further mitigating excessively long generation on top of existing precision allocations. In supplementary comparisons at similar storage budgets, window maintenance also achieves higher macro-averages and shorter mean outputs, demonstrating the value of devoting part of the storage capacity to recent-information maintenance.

\paragraph{The mechanisms behind excessively long outputs and repetitive generation require further study.}
PW mitigates verbose outputs after quantization, but some responses still reach the token limit or terminate because of repetition. We have not yet fully characterized how information written by tokens is retained and read during state recurrences, or how quantization perturbations affect its subsequent use. Future work will further analyze the contribution of recent writes to predictions and investigate its relationship to generation feedback, stalled reasoning, and repetitive outputs.

\paragraph{PW characterization and algorithm design remain open to improvement.}
Our current evaluation mainly considers a fixed window configuration. Systematic experiments on properties such as window length, queue precision, and branch-merging strategies remain incomplete. We will extend the corresponding comparisons and ablations, improve the window-maintenance algorithm, and further investigate joint optimization of window maintenance and global precision allocation.

\paragraph{Dedicated efficient inference kernels remain to be developed.}
We have not yet developed dedicated efficient inference kernels for PW. Future work will develop and optimize kernels for propagating and reading recent branches and merging them into the historical state, reduce computation and memory-access overhead in queue maintenance, and evaluate end-to-end inference latency and throughput.

\paragraph{Validation covers a limited set of models and methods.}
The current experiments mainly use Qwen3.5-4B/9B, with limited coverage of other model sizes, model families, and recurrent-state quantization methods. We will expand the range of models and comparison methods and assess performance across architectures and deployment conditions using task accuracy, generation length, and inference-performance metrics.

\section{Conclusion}
\label{sec:conclusion}
\suppressfloats[t]

We propose GGA and PW, which use writeback sensitivity to future prediction loss to guide precision allocation across heads and maintain independent floating-point branches according to write age, respectively. At @$4$ on Qwen3.5-4B/9B, GGA allocation alone improves scores by $19.48$/$29.84$ percentage points; adding PW-S8 yields task macro-averages of $69.83\%$/$78.63\%$, improving over STEPQuant@$4$ by $38.39$/$41.21$ percentage points. At @$6$, STEPQuant substantially recovers task performance, while GGA still reduces mean generation length, averaged equally across tasks, by approximately $10.45\%$/$14.65\%$ with similar or higher accuracy. At similar storage budgets, GGA@$4$+PW-S8 further reduces mean output length by $9.91\%$/$10.80\%$ relative to GGA@$5$, while increasing the macro-average by $2.61$/$0.41$ percentage points. Allocation diagnostics and equivalent-bit accounting further characterize precision placement and window storage overhead, respectively. Future work will expand sampling and investigate additional model families and window lengths.

\FloatBarrier
\appendix
\makeatletter
\setlength{\@fptop}{0pt}
\setlength{\@fpsep}{12pt}
\setlength{\@fpbot}{0pt plus 1fil}
\setlength{\@dblfptop}{0pt}
\setlength{\@dblfpsep}{12pt}
\setlength{\@dblfpbot}{0pt plus 1fil}
\makeatother

\section{Equivalence of PW under Exact Arithmetic}
\label{app:equivalence}
\suppressfloats[t]

Assume that the initial old state equals the initial GDN state and that the recent queue is empty. The two recurrences use identical exogenous $q,k,v,\beta,\gamma$, all operations use infinite precision, and the old state is not quantized. Denote the sum of the two components by
\begin{equation}
 S_t^{\mathrm{PW}}=O_t+\sum_{j\in\mathcal J_t}p_{j,t}v_j^{\top}.
\end{equation}
The initial conditions give $S_0^{\mathrm{PW}}=S_0$. If this equality holds at the preceding step, applying $A_t$ to the old state and all existing left factors and then appending $B_t$ yields
\begin{align}
 S_t^{\mathrm{PW}}
 &=A_tO_{t-1}+\sum_{j\in\mathcal J_{t-1}}A_tp_{j,t-1}v_j^{\top}+B_t\notag\\
 &=A_tS_{t-1}^{\mathrm{PW}}+B_t=S_t.
\end{align}
When the window is full, eviction only moves the same term $p_{t-m,t}v_{t-m}^{\top}$ from recent to old, leaving their sum unchanged. Induction therefore establishes that the logical state and current readout are identical at every step. Finite-precision rounding, quantization feedback, and activation changes during free generation fall outside the assumptions of this equivalence.

With quantization, $\Q(O_t)+H_t$ generally differs from $\Q(O_t+H_t)$, where $H_t$ is the sum of the recent branches. PW thus changes the state representation and the location at which errors are injected. The algebraic equivalence above does not guarantee that elementwise errors necessarily decrease in finite-precision deployment, nor that free-generation trajectories match those of the high-precision model.

\FloatBarrier

\section{Calibration, Deployment, and Storage Accounting}
\label{app:implementation}
\suppressfloats[t]

\subsection{Frozen Plans and Budget Settings}
Each evaluated model contains $24\times32=768$ state/value heads of size $128\times128$, and each plan selects 32 FP16 pivots. Under the @$4$, @$5$, and @$6$ settings, the integer bit-width sum budgets $B_{\mathrm{int}}$ for the remaining 736 heads are 2816, 3584, and 4352, respectively. The @$5$ setting increases the budget using the same calibration costs and the same candidate formats as @$4$. Table~\ref{tab:plan-histogram} lists the bit-width distributions of the frozen plans; GGA and STEP may select different pivot identities.

\begin{table}[!htbp]
\centering
\fontsize{8.5}{10}\selectfont
\setlength{\tabcolsep}{3pt}
\caption{Head counts in the precision allocation plans. PW configurations not listed separately use the same bit-width tables as their corresponding configurations without a window.}
\label{tab:plan-histogram}
\begin{tabular*}{\textwidth}{@{\extracolsep{\fill}}llrrrrr@{}}
\toprule
Model & Allocation & INT2 & INT4 & INT6 & INT8 & FP16\\
\midrule
4B & STEP@$4$ & 380 & 119 & 158 & 79 & 32\\
4B & GGA@$4$ & 369 & 139 & 151 & 77 & 32\\
4B & STEP@$5$ & 214 & 185 & 140 & 197 & 32\\
4B & GGA@$5$ & 215 & 171 & 165 & 185 & 32\\
4B & STEP@$6$ & 0 & 302 & 164 & 270 & 32\\
4B & GGA@$6$ & 0 & 293 & 182 & 261 & 32\\
4B & GGA@$6$+PW-S8 & 0 & 319 & 130 & 287 & 32\\
\midrule
9B & STEP@$4$ & 372 & 132 & 156 & 76 & 32\\
9B & GGA@$4$ & 361 & 167 & 119 & 89 & 32\\
9B & STEP@$5$ & 213 & 185 & 143 & 195 & 32\\
9B & GGA@$5$ & 218 & 157 & 184 & 177 & 32\\
9B & STEP@$6$ & 0 & 299 & 170 & 267 & 32\\
9B & GGA@$6$ & 0 & 285 & 198 & 253 & 32\\
9B & GGA@$6$+PW-S8 & 0 & 315 & 138 & 283 & 32\\
\bottomrule
\end{tabular*}
\end{table}

\subsection{Solving the Global Allocation by Dynamic Programming}
\label{app:allocation-dp}

To solve Eq.~\eqref{eq:allocation}, denote the non-pivot heads in a fixed order by $u_1,\ldots,u_n$. Let $F(j,s)$ be the minimum cost for the first $j$ heads when their bit widths sum to exactly $s$. The recurrence is
\begin{equation}
 F(j,s)=\min_{\substack{b\in\mathcal B\\b\le s}}
 \bigl[F(j-1,s-b)+C_{u_j,b}^{\mathrm{GGA}}\bigr].
 \label{eq:dp}
\end{equation}
The boundary condition is $F(0,0)=0$, with all other initial values set to $+\infty$. Backtracking from $F(n,B_{\mathrm{int}})$ gives the integer bit-width table.

The preliminary allocation applies the same recurrence to all heads, with a target budget of $\bar b|\U|$.

\subsection{Calibration and Deployment Settings}
Calibration for standalone GGA allocation uses 32 frozen token sequences from the WikiText-2 training split, each containing 2048 tokens, with data sampling seed 0. Each sequence is processed in a separate forward pass, and the language-model loss is the sum of the 2047 next-token cross-entropies within that sequence. Updated states are collected every 8 tokens, giving 8192 candidate positions per layer, from which reservoir sampling with seed 0 retains 64 snapshots. State residuals and writeback adjoints are paired at the same sequence and token positions, and Eq.~\eqref{eq:gga-cost} is averaged over these snapshots. Gradients are taken from the original GDN head readouts, and the adjoints are computed using reference activations that have been normalized, scaled, and mapped to the state heads.

The implementation divides all GGA costs across the model by the median positive cost. This common positive rescaling changes neither the optimal solution of the objective nor the pivot risk ranking; it is not normalization per layer or per head. Candidate costs and precision allocations for each model are determined offline on the calibration set. Inference uses the frozen bit widths and pivots for each configuration. The local diagnostics use the corresponding STEP/GGA@$4$ allocation configurations.

Calibration for GGA@$6$+PW-S8 uses a PW reference trajectory with window length $m=8$, in which the historical states of non-pivot heads are stored in FP32 and the recent queue uses a scaled FP8 representation. Candidate costs retain the squared-gradient weighting in Eq.~\eqref{eq:gga-cost}, using snapshots of the historical state $O_{u,t}$ and the gradient of the full-model cross-entropy loss with respect to that writeback state as the sensitivity. Backpropagation through FP8 and FP16 rounding uses an identity straight-through estimator (identity STE). The 32 FP16 pivots of the corresponding GGA@$6$ configuration remain fixed; global dynamic programming allocates the remaining heads over the candidate set $\{4,6,8\}$ with an integer bit-width sum of 4352.

\subsection{FP8 Encoding and Online Updates for PW}
\label{app:pw-implementation}

The recent queue in PW uses E4M3FN FP8 encoding. Each $p/v$ vector of each branch in each head stores its own FP32 scale. Propagation, readout, and eviction use the reconstructed FP32 values. PW-S8 denotes a window length of $m=8$. For a nonzero vector $x$, define
\begin{equation}
 s(x)=\frac{\|x\|_{\infty}}{448},\qquad
 \Rfp(x)=s(x)\,\mathrm{FP32}\!\left[
 \mathrm{FP8}_{\mathrm{E4M3FN}}\!\left(\frac{x}{s(x)}\right)\right],
 \label{eq:fp8-map}
\end{equation}
with $\Rfp(0)=0$. The map $\Rfp$ composes encoding and reconstruction; persistent storage retains only the FP8 codes and the scale. The finite-precision recent branches are
\begin{equation}
 \begin{aligned}
 \bar p_{j,j}&=\Rfp(\beta_jk_j),\qquad
 \bar v_j=\Rfp(v_j),\\
 \bar p_{j,t}&=\Rfp(A_t\bar p_{j,t-1}).
 \end{aligned}
 \label{eq:fp8}
\end{equation}
After propagation at each step, existing left factors are re-encoded and their scales updated; their reconstructed values are then used for the current readout and eviction. Each $\bar v_j$ remains fixed after enqueueing. The current new write contributes to the current readout using FP32 intermediate values, and its $p/v$ vectors are encoded and enqueued only afterward.

Algorithm~\ref{alg:pw} represents recent storage as a bounded queue $\mathcal R$ ordered by birth time. Queue elements are written as their FP32 reconstructed values $\bar p,\bar v$. Persistent matrices and queue vectors are decoded before computation. The map $\Rfp$ implicitly includes FP8 encoding, scale writeback, and reconstruction, while $\Q_b$ denotes the reconstructed value after old-state encoding. In the algorithm, $A$ is implemented using the low-rank form in Eq.~\eqref{eq:p-update}, without constructing a dense matrix. FP32 intermediate values are used only within the current step.

\begin{algorithm}[t]
\caption{PW decode update for one head}
\label{alg:pw}
\small
\begin{algorithmic}[1]
\Require Reconstructed persistent matrix $\hat O$, FP8 queue $\mathcal R$ with scales, current $q,k,v,\gamma,\beta$
\Statex \hspace{\algorithmicindent} Frozen format $b$ or FP16 pivot, window $m\ge1$, $|\mathcal R|\le m$
\Ensure Current readout $o$, persistent representation and queue for the next step
\State $A\gets\gamma(I-\beta kk^{\top})$, $p_{\mathrm{new}}\gets\beta k$
\If{the current head is an FP16 pivot}
  \State $S^{\mathrm{cand}}\gets A\hat O+p_{\mathrm{new}}v^{\top}$
  \State $o\gets(S^{\mathrm{cand}})^{\top}q$
  \State $\hat O\gets\mathrm{FP16}(S^{\mathrm{cand}})$
  \State \Return $o,\hat O,\varnothing$
\EndIf
\State $O^{\mathrm{pre}}\gets A\hat O$
\For{each existing vector pair $(\bar p,\bar v)$ in the queue}
  \State $\bar p\gets\Rfp(A\bar p)$ \Comment{Decode and propagate; re-encode and update the scale}
\EndFor
\State $o\gets(O^{\mathrm{pre}})^{\top}q+\displaystyle\sum_{(\bar p,\bar v)\in\mathcal R}\bar v(\bar p^{\top}q)+v(p_{\mathrm{new}}^{\top}q)$
\State $O^{\mathrm{merge}}\gets O^{\mathrm{pre}}$
\If{$|\mathcal R|=m$}
  \State $(\bar p_{\mathrm{old}},\bar v_{\mathrm{old}})\gets$ \Call{PopOldest}{$\mathcal R$}
  \State $O^{\mathrm{merge}}\gets O^{\mathrm{merge}}+\bar p_{\mathrm{old}}\bar v_{\mathrm{old}}^{\top}$
\EndIf
\State Append $(\Rfp(p_{\mathrm{new}}),\Rfp(v))$ to $\mathcal R$
\State $\hat O\gets\Q_b(O^{\mathrm{merge}})$ \Comment{Encode and write back old at every step}
\State \Return $o,\hat O,\mathcal R$
\end{algorithmic}
\end{algorithm}

The oldest branch is evicted only after the current readout is complete. The current new write contributes to the readout as temporary vectors and is then enqueued, so no additional persistent queue slot numbered $m+1$ is required.

FP16 pivots retain full-state updates and an FP16 persistent representation, with no PW queue enabled. Prefill follows the base execution path, and the last $m$ writes of the prompt are not automatically added to the decode queue. Queue storage accounting is given in Appendix~\ref{app:storage}.

Propagation of the recent left factors costs $O(md_k)$, and recent readout costs $O(m(d_k+d_v))$. The old-state update, the outer product of the evicted branch, and encoding also involve $O(d_kd_v)$ operations and quantization encoding overhead. Combined configurations and equivalent state bit widths are given in Appendix~\ref{app:storage}.

\subsection{State Storage Accounting and Queue Capacity}
\label{app:storage}

In the combined configuration, GGA determines the bit-width table and pivot set offline, and PW maintains the old state online at the corresponding bit widths. Deployment involves no backpropagation, and both the window length and the bit width of every head are frozen.

This section accounts for the persistent data arrays of a single request's state slot, converting storage to equivalent bit/value using the total number of elements in all GDN state matrices, \mbox{$N_S=\sum_{u\in\U}d_{k,u}d_{v,u}$}, as the denominator. The accounting includes integer codes, quantization scales, FP16 pivots, and reserved PW queues. It excludes indices, counters, shared plan metadata, unused request slots in the state pool, and temporary computation workspace, as well as weights, full-attention KV caches, and convolution caches.

Let $\U_I=\U\setminus\Piv$ be the set of non-pivot heads, and let $M_{Q,u}(b_u)$ be the packed old-state storage in bytes for an integer head, including integer codes and the scales for that format. Let $\U_R$ be the set of heads for which PW queues are actually reserved. Each FP8 branch stores two encoded vectors and their respective FP32 scales, requiring $d_{k,u}+d_{v,u}+8$ bytes. The total number of bytes in the persistent data arrays being accounted for is
\begin{equation}
 M_{\mathrm{persist}}=\sum_{u\in\U_I}M_{Q,u}(b_u)
 +\sum_{u\in\Piv}2d_{k,u}d_{v,u}
 +m\sum_{u\in\U_R}(d_{k,u}+d_{v,u}+8).
 \label{eq:storage}
\end{equation}
The corresponding equivalent state bit width and additional queue bit width are
\begin{equation}
 b_{\mathrm{eff}}=\frac{8M_{\mathrm{persist}}}{N_S},
 \qquad
 \Delta b_{\mathrm{queue}}
 =\frac{8m}{N_S}\sum_{u\in\U_R}(d_{k,u}+d_{v,u}+8).
 \label{eq:effective-bits}
\end{equation}

When PW is disabled, $\U_R=\varnothing$. With PW enabled, the current layout reserves queues for all heads, so $\U_R=\U$.

For $d_k=d_v=128,m=8$, each FP8 vector occupies $128+4=132$ bytes, including its codes and FP32 scale. Each head with a queue requires $8\times2\times132=2112$ bytes, equivalent to 1.03125 bit/value of that head's state matrix. Allocating queues only for the 736 non-pivot heads would add approximately 0.98828 bit/value when averaged over all 768 heads. The current storage layout reserves queues for every head, including pivots; Table~\ref{tab:storage} therefore accounts for an additional 1.03125 bit/value. Small metadata such as counters is accounted for separately.

Table~\ref{tab:storage} converts the nominal budgets to storage under the same accounting convention: @$5$ requires approximately $5.68$ bit/value, and @$4$+PW-S8 approximately $5.78$--$5.79$ bit/value; @$6$ and @$6$+PW-S8 require $6.573$ and $7.604$ bit/value, respectively.

\begin{table}[!htbp]
\centering
\fontsize{8.5}{10}\selectfont
\setlength{\tabcolsep}{3pt}
\caption{Equivalent state bit widths (bit/value) for different allocation and window settings. State encoding, scales, FP16 pivots, and PW queues are included using the fixed array layout. FP32/FP16 floating-point states require 32/16 bit/value, respectively.}
\label{tab:storage}
\begin{tabular*}{\textwidth}{@{\extracolsep{\fill}}llrrrrr@{}}
\toprule
Model & Allocation & @$4$ & @$4$+PW & @$5$ & @$6$ & @$6$+PW\\
\midrule
4B & STEP & 4.758 & 5.790 & 5.677 & 6.573 & 7.604\\
 & GGA & 4.753 & 5.784 & 5.678 & 6.573 & 7.604\\
\midrule
9B & STEP & 4.755 & 5.786 & 5.677 & 6.573 & 7.604\\
 & GGA & 4.749 & 5.780 & 5.679 & 6.573 & 7.604\\
\bottomrule
\end{tabular*}
\end{table}

\FloatBarrier

\section{Protocol and Binned Data for Precision Allocation Diagnostics}
\label{app:allocation-diagnostic}
\suppressfloats[t]

This section supplements the precision configuration analysis in Section~\ref{sec:allocation-observation} with the experimental protocol, metric definitions, and binned data. The analysis uses all 768 state heads of Qwen3.5-4B.

\subsection{Reference Trajectories and Single Interventions}
This analysis uses the H256 local diagnostic set. Questions are drawn from the evaluation tasks listed in the main text and selected according to the frozen source fingerprint and hash order, excluding calibration and previously used development questions. Selection does not depend on differences between methods or answer correctness. The set must satisfy the replay length requirement and is used for local intervention analysis; results are averaged equally over questions rather than datasets.

Given a generated sequence $g[0:T]$, at the after-call-$d$ point the model has consumed $g[0:d]$, and the logits that currently predict $g[d]$ have already been determined. We then quantize the entire accumulated state of only the target head once, reconstruct it in FP32, and replay the fixed sequence $g[d:d+256]$. The two checkpoints are determined by the trajectory length:
\begin{align}
 D_{\max}&=T-256-1,\\
 d_1&=1+\lfloor0.25(D_{\max}-1)\rfloor,\\
 d_2&=\max\{d_1+1,\;1+\lfloor0.65(D_{\max}-1)\rfloor\}.
\end{align}
The current logits, which have already been produced, are excluded from the 256 steps after the intervention. Teacher forcing fixes the token inputs but does not fix the activations or states of the whole model after the intervention. This therefore differs from the fixed exogenous activation condition used for the algebraic derivations in the Method section.

Every head is tested at INT2, INT4, INT6, and INT8, using the frozen candidate quantizer with dual-axis fitting, row-impact, INT2 group32, and FP16 scales.

\subsection{Empirical Quantization Impact and Binned Statistics}
For question $i$, checkpoint $p$, head $u$, and integer format $b\in\{2,4,6,8\}$, define the mean predictive distribution divergence over the subsequent 256 steps as
\begin{equation}
 R_{i,p,u,b}=\frac{1}{256}\sum_{s=1}^{256}
 D_{\mathrm{KL}}\!\left(P^{\mathrm{FP}}_{i,p,s}\,\Vert\,
 P^{\mathrm{inj}}_{i,p,u,b,s}\right),
 \label{eq:diagnostic-kl}
\end{equation}
where $P^{\mathrm{FP}}$ and $P^{\mathrm{inj}}$ denote the model's predictive probability distributions over the vocabulary along the reference trajectory and after a single intervention, respectively. Let $N_{\mathrm{diag}}$ be the number of diagnostic questions. We first average the two checkpoints within each question and then average equally over questions, and define a composite quantization impact across the four precisions:
\begin{equation}
 \begin{aligned}
 R_{u,b}&=\frac{1}{N_{\mathrm{diag}}}\sum_{i=1}^{N_{\mathrm{diag}}}\frac{1}{2}\sum_{p=1}^{2}R_{i,p,u,b},\\
 I_u&=\frac{1}{4}\sum_{b\in\{2,4,6,8\}}R_{u,b}.
 \end{aligned}
 \label{eq:joint-impact}
\end{equation}
The metric $I_u$ summarizes quantization residuals and their subsequent propagation effects under common candidate precisions, providing the same observational coordinate for the two precision configurations. GGA solves for bit widths using the calibration costs in Eq.~\eqref{eq:gga-cost}.

Heads are partitioned by $I_u$ into 11 equal-width bins of width $0.30\times10^{-4}$. For each nonempty bin $\mathcal I_k$ ($n_k>0$), both methods compute the mean state-value bit width over the same set of heads:
\begin{equation}
 \bar b_{a,k}=\frac{1}{n_k}\sum_{u:I_u\in\mathcal I_k}b_u^{(a)},
 \qquad n_k=\bigl|\{u:I_u\in\mathcal I_k\}\bigr|,
 \label{eq:binned-precision}
\end{equation}
where $a\in\{\mathrm{STEP},\mathrm{GGA}\}$. Integer formats count according to their bit widths, and FP16 counts as 16 bits. Both configurations contain 32 FP16 pivots, and the bit widths of the 736 integer heads sum to 2816 in each configuration, giving a weighted mean of $4.33$ bit/value across all heads. This statistic accounts for state-value encoding; the storage overhead of scales and metadata is accounted for separately in Appendix~\ref{app:storage}.

The bin boundaries are $[0.30,0.60,\ldots,3.60]\times10^{-4}$. Bins are left-closed and right-open, except for the final bin, which is closed on the right. All 768 heads are included, and empty bins are not filled with zeros. Table~\ref{tab:impact-bins} reports the final statistics for all 11 bins, with displayed values rounded to two decimal places. The composite metric and this binning are descriptive analyses conducted after the experiment; the source table has already been aggregated across questions and checkpoints.

Both configurations query the same table of measured impacts. For the 61 heads in the union of the FP16 pivot sets, we additionally measure $\mathrm{FP32}(\mathrm{FP16}(S))$. The same definition of $R$ extends to these additionally tested FP16 formats, whereas the composite metric $I_u$ still averages only the four integer formats. Querying $R_{u,b_u}$ at each head's assigned bit width and averaging equally over heads gives mean single-head local KL values of $2.15559\times10^{-4}$ for STEPQuant and $2.16013\times10^{-4}$ for GGA, with GGA approximately $0.21\%$ higher. This statistic corresponds to a single intervention on each head; the joint effects of persistent quantization are examined separately through task evaluation.

\begin{table}[!htbp]
\centering
\begin{minipage}{.82\linewidth}
\centering
\fontsize{8.5}{10}\selectfont
\setlength{\tabcolsep}{3pt}
\caption{Mean state-value bit width under the common composite impact metric. Bin units are $10^{-4}$, and FP16 counts as 16 bits.}
\label{tab:impact-bins}
\begin{tabular*}{\linewidth}{@{\extracolsep{\fill}}lrrr@{}}
\toprule
Composite impact bin & Heads & STEPQuant & GGA\\
\midrule
$[0.30,0.60)$ & 10 & 2.00 & 2.00\\
$[0.60,0.90)$ & 7 & 4.00 & 2.57\\
$[0.90,1.20)$ & 25 & 2.96 & 2.40\\
$[1.20,1.50)$ & 56 & 5.14 & 3.36\\
$[1.50,1.80)$ & 81 & 5.28 & 3.09\\
$[1.80,2.10)$ & 104 & 4.37 & 3.13\\
$[2.10,2.40)$ & 127 & 4.19 & 4.02\\
$[2.40,2.70)$ & 231 & 3.73 & 4.56\\
$[2.70,3.00)$ & 125 & 5.02 & 6.96\\
$[3.00,3.30)$ & 0 & -- & --\\
$[3.30,3.60]$ & 2 & 7.00 & 16.00\\
\bottomrule
\end{tabular*}
\end{minipage}
\end{table}

\FloatBarrier

\section{Evaluation Protocol and Supplementary Statistics}
\label{app:evaluation-protocol}
\suppressfloats[t]

The random evaluation subsets for all tasks are fixed before inference, and models and methods share the same question lists. LiveCodeBench v6 questions are selected using a predetermined hash ordering, with question identifiers, source file hashes, and the original ordering frozen. The frozen upstream evaluation code and grading package versions are listed in Table~\ref{tab:grader-revisions}. For IFBench, this paper uses prompt strict rather than the prompt loose metric usually reported in its paper.

\begin{table}[!htbp]
\centering
\begin{minipage}{.82\linewidth}
\centering
\fontsize{8.5}{10}\selectfont
\setlength{\tabcolsep}{3pt}
\caption{Frozen upstream code and grading dependencies for the formal evaluation. Code repositories are identified by the first 12 characters of the commit hash, and grading packages by their version numbers.}
\label{tab:grader-revisions}
\begin{tabular*}{\linewidth}{@{\extracolsep{\fill}}lll@{}}
\toprule
Implementation & Task & Version\\
\midrule
MathArena & AIME, HMMT & \texttt{b89f2f0ad64c}\\
math-verify & MATH500 & \texttt{0.9.0}\\
GPQA & GPQA Diamond & \texttt{56686c06f5e1}\\
EvalPlus & HumanEval+ & \texttt{26d6d00bb1fd}\\
IFBench & IFBench & \texttt{1c40f0c10d9b}\\
LiveCodeBench & LCB v6 subset & \texttt{28fef95ea8c9}\\
\bottomrule
\end{tabular*}
\end{minipage}
\end{table}

In addition to natural EOS and the 100K length cap, all settings use the same stopping rule for short-period repetition. Repetition stops and length-cap terminations are reported separately. The final answer is the content after the last \texttt{</think>}; it is counted as missing if empty after removing termination markers. Missing final answers can overlap with any stopping category, and all responses remain in the evaluation denominator.

Tables~\ref{tab:task-lengths} and~\ref{tab:budget-comparison} provide the complete task-level lengths and numeric results for the similar-budget comparison in Section~\ref{sec:budget-comparison}.

\begin{table}[!htbp]
\centering
\fontsize{8.5}{10}\selectfont
\setlength{\tabcolsep}{3pt}
\caption{Mean full output lengths (thousand tokens) by task for 4B/9B at @$4$ and @$6$, alongside floating-point state references. Lengths include thinking and final, exclude the input prompt, and retain all stopping types. Mean is the equally weighted average of the seven task means, computed from unrounded values. Task abbreviations follow Table~\ref{tab:main-results}.}
\label{tab:task-lengths}
\begin{tabular*}{\textwidth}{@{\extracolsep{\fill}}llrrrrrrrr@{}}
\toprule
Model & Method & AIME & HMMT & GPQA & HE+ & MATH & IFB & LCB & Mean\\
\midrule
4B & FP32 & 39.79 & 53.10 & 26.62 & 1.27 & 15.06 & 17.81 & 20.94 & 24.94\\
 & FP16 & 47.98 & 63.50 & 28.41 & 1.30 & 16.46 & 20.92 & 22.48 & 28.72\\
\cmidrule(lr){2-10}
 & Naive INT4 & 3.74 & 3.33 & 3.03 & 2.18 & 3.18 & 2.87 & 2.64 & 2.99\\
 & STEP@$4$ & 94.21 & 97.13 & 84.17 & 2.06 & 56.67 & 27.30 & 46.06 & 58.23\\
 & GGA@$4$ & 81.36 & 86.87 & 48.60 & 1.41 & 30.33 & 16.66 & 35.46 & 42.96\\
 & STEP@$4$+PW-S8 & 70.31 & 81.62 & 34.76 & 1.44 & 19.68 & 12.85 & 25.38 & 35.15\\
 & GGA@$4$+PW-S8 & 54.13 & 67.48 & 30.22 & 1.26 & 15.06 & 11.25 & 24.10 & 29.07\\
\cmidrule(lr){2-10}
 & Naive INT6 & 32.21 & 31.65 & 21.74 & 7.53 & 23.09 & 7.11 & 18.35 & 20.24\\
 & STEP@$6$ & 59.40 & 74.65 & 34.66 & 1.29 & 17.19 & 16.59 & 24.73 & 32.64\\
 & GGA@$6$ & 52.69 & 66.17 & 30.44 & 1.20 & 14.93 & 15.90 & 23.31 & 29.23\\
 & STEP@$6$+PW-S8 & 40.14 & 52.34 & 20.96 & 1.23 & 11.18 & 11.81 & 16.41 & 22.01\\
 & GGA@$6$+PW-S8 & 44.63 & 59.15 & 25.25 & 1.30 & 12.66 & 11.64 & 19.36 & 24.86\\
\midrule
9B & FP32 & 36.05 & 48.40 & 24.72 & 1.01 & 12.67 & 14.82 & 21.06 & 22.68\\
 & FP16 & 40.29 & 54.15 & 26.16 & 1.03 & 14.13 & 15.68 & 22.04 & 24.78\\
\cmidrule(lr){2-10}
 & Naive INT4 & 3.75 & 3.67 & 3.82 & 2.74 & 3.76 & 3.12 & 3.40 & 3.46\\
 & STEP@$4$ & 93.51 & 94.62 & 77.66 & 1.31 & 48.28 & 27.54 & 45.02 & 55.42\\
 & GGA@$4$ & 69.27 & 77.76 & 42.69 & 1.25 & 22.36 & 16.46 & 34.68 & 37.78\\
 & STEP@$4$+PW-S8 & 60.25 & 71.16 & 35.54 & 1.12 & 16.20 & 13.97 & 28.47 & 32.39\\
 & GGA@$4$+PW-S8 & 43.23 & 53.68 & 24.53 & 1.05 & 11.78 & 10.71 & 22.24 & 23.89\\
\cmidrule(lr){2-10}
 & Naive INT6 & 18.16 & 18.69 & 13.56 & 3.79 & 13.29 & 7.50 & 9.30 & 12.04\\
 & STEP@$6$ & 51.83 & 64.53 & 33.67 & 1.04 & 14.00 & 15.71 & 24.67 & 29.35\\
 & GGA@$6$ & 43.74 & 57.42 & 26.73 & 1.02 & 11.79 & 12.76 & 21.90 & 25.05\\
 & STEP@$6$+PW-S8 & 37.03 & 51.47 & 21.98 & 1.05 & 10.18 & 13.76 & 19.01 & 22.07\\
 & GGA@$6$+PW-S8 & 38.04 & 51.75 & 22.14 & 1.03 & 10.50 & 10.65 & 18.87 & 21.85\\
\bottomrule
\end{tabular*}
\end{table}

\begin{table}[!htbp]
\centering
\fontsize{8.5}{10}\selectfont
\setlength{\tabcolsep}{3pt}
\caption{Paired comparisons at similar storage budgets. Mean length weights tasks equally and is measured in thousand tokens; scores are seven-task macro-averages (\%). Boldface indicates the higher macro-average and shorter mean output within each allocation method.}
\label{tab:budget-comparison}
\begin{tabular*}{\textwidth}{@{\extracolsep{\fill}}llrrr@{}}
\toprule
Model & Setting & bit/value & Macro-avg. & Mean length\\
\midrule
4B & STEP@$5$ & 5.677 & 60.38 & 37.25\\
 & STEP@$4$+PW-S8 & 5.790 & \textbf{63.67} & \textbf{35.15}\\
\cmidrule(lr){2-5}
 & GGA@$5$ & 5.678 & 67.23 & 32.27\\
 & GGA@$4$+PW-S8 & 5.784 & \textbf{69.83} & \textbf{29.07}\\
\midrule
9B & STEP@$5$ & 5.677 & 68.46 & 35.03\\
 & STEP@$4$+PW-S8 & 5.786 & \textbf{70.02} & \textbf{32.39}\\
\cmidrule(lr){2-5}
 & GGA@$5$ & 5.679 & 78.22 & 26.78\\
 & GGA@$4$+PW-S8 & 5.780 & \textbf{78.63} & \textbf{23.89}\\
\bottomrule
\end{tabular*}
\end{table}

\begin{figure}[!htbp]
\centering
\begin{minipage}[t]{.48\linewidth}
\centering
\includegraphics[width=66mm]{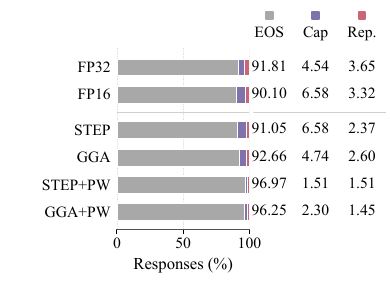}
\caption{Stopping categories on Qwen3.5-4B at @$6$, alongside floating-point state references.}
\label{fig:termination-4b-6}
\end{minipage}
\hfill
\begin{minipage}[t]{.48\linewidth}
\centering
\includegraphics[width=66mm]{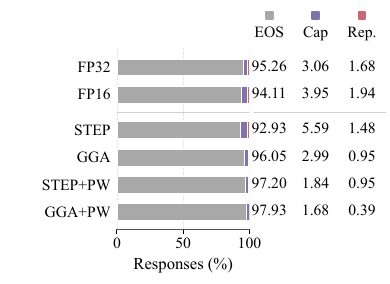}
\caption{Stopping categories on Qwen3.5-9B at @$6$, alongside floating-point state references.}
\label{fig:termination-9b-6}
\end{minipage}
\end{figure}

Figures~\ref{fig:termination-4b-6} and~\ref{fig:termination-9b-6} supplement the main-text stopping-category plots at @$6$. Tables~\ref{tab:termination} and~\ref{tab:supplementary-termination} report termination statistics for the floating-point references and the @$4$ and @$6$ settings, and for the supplementary @$5$ settings, respectively. Stopping-category percentages are aggregated over responses, while mean lengths retain the equal task weighting used in the main text. Table~\ref{tab:budget-task-lengths} additionally reports per-task mean lengths for @$5$ and @$4$+PW-S8, corresponding to the similar-budget comparison in Section~\ref{sec:budget-comparison}.

\begin{table}[!htbp]
\centering
\fontsize{8.5}{10}\selectfont
\setlength{\tabcolsep}{3pt}
\caption{Generation lengths and termination statistics for the floating-point references and the @$4$ and @$6$ settings. Percentages are aggregated over responses. Mean lengths are in thousands of tokens and are the equally weighted mean of each task's mean full output length, including thinking and excluding input prompts. EOS, length-cap termination, and repetition stops are mutually exclusive; missing final answers may overlap with them.}
\label{tab:termination}
\begin{tabular*}{\textwidth}{@{\extracolsep{\fill}}llrrrrr@{}}
\toprule
Model & Method & EOS & 100K cap & Repetition & No final & Mean length\\
\midrule
4B & FP32 & 91.81 & 4.54 & 3.65 & 7.54 & 24.94\\
 & FP16 & 90.10 & 6.58 & 3.32 & 9.38 & 28.72\\
\cmidrule(lr){2-7}
 & Naive INT4 & 100.00 & 0.00 & 0.00 & 11.62 & 2.99\\
 & STEP@$4$ & 53.37 & 36.66 & 9.97 & 45.74 & 58.23\\
 & GGA@$4$ & 79.43 & 14.18 & 6.38 & 19.55 & 42.96\\
 & STEP@$4$+PW-S8 & 91.12 & 8.36 & 0.53 & 8.88 & 35.15\\
 & GGA@$4$+PW-S8 & 94.93 & 4.61 & 0.46 & 5.30 & 29.07\\
\cmidrule(lr){2-7}
 & Naive INT6 & 92.27 & 0.03 & 7.70 & 13.62 & 20.24\\
 & STEP@$6$ & 91.05 & 6.58 & 2.37 & 8.75 & 32.64\\
 & GGA@$6$ & 92.66 & 4.74 & 2.60 & 7.17 & 29.23\\
 & STEP@$6$+PW-S8 & 96.97 & 1.51 & 1.51 & 3.29 & 22.01\\
 & GGA@$6$+PW-S8 & 96.25 & 2.30 & 1.45 & 3.82 & 24.86\\
\midrule
9B & FP32 & 95.26 & 3.06 & 1.68 & 4.28 & 22.68\\
 & FP16 & 94.11 & 3.95 & 1.94 & 5.56 & 24.78\\
\cmidrule(lr){2-7}
 & Naive INT4 & 100.00 & 0.00 & 0.00 & 15.76 & 3.46\\
 & STEP@$4$ & 59.62 & 30.67 & 9.71 & 39.59 & 55.42\\
 & GGA@$4$ & 90.69 & 8.23 & 1.09 & 8.49 & 37.78\\
 & STEP@$4$+PW-S8 & 91.67 & 7.01 & 1.32 & 8.46 & 32.39\\
 & GGA@$4$+PW-S8 & 97.47 & 2.44 & 0.10 & 2.83 & 23.89\\
\cmidrule(lr){2-7}
 & Naive INT6 & 99.11 & 0.00 & 0.89 & 11.85 & 12.04\\
 & STEP@$6$ & 92.93 & 5.59 & 1.48 & 6.81 & 29.35\\
 & GGA@$6$ & 96.05 & 2.99 & 0.95 & 3.75 & 25.05\\
 & STEP@$6$+PW-S8 & 97.20 & 1.84 & 0.95 & 2.86 & 22.07\\
 & GGA@$6$+PW-S8 & 97.93 & 1.68 & 0.39 & 2.17 & 21.85\\
\bottomrule
\end{tabular*}
\end{table}

\begin{table}[!htbp]
\centering
\fontsize{8.5}{10}\selectfont
\setlength{\tabcolsep}{3pt}
\caption{Generation lengths and termination statistics for the supplementary @$5$ configurations. Percentages are in \%, and mean lengths are in thousands of tokens; the accounting convention is the same as in Table~\ref{tab:termination}.}
\label{tab:supplementary-termination}
\begin{tabular*}{\textwidth}{@{\extracolsep{\fill}}llrrrrr@{}}
\toprule
Model & Method & EOS & 100K cap & Repetition & No final & Mean length\\
\midrule
4B & STEP@$5$ & 83.38 & 9.64 & 6.98 & 16.42 & 37.25\\
 & GGA@$5$ & 89.87 & 5.03 & 5.10 & 9.94 & 32.27\\
\midrule
9B & STEP@$5$ & 89.40 & 8.03 & 2.57 & 10.37 & 35.03\\
 & GGA@$5$ & 95.56 & 3.29 & 1.15 & 4.18 & 26.78\\
\bottomrule
\end{tabular*}
\end{table}

\clearpage

\begin{table}[!htbp]
\centering
\fontsize{8.5}{10}\selectfont
\setlength{\tabcolsep}{3pt}
\caption{Mean full output lengths for each task in the supplementary similar-budget comparison (thousands of tokens). The @$5$ and @$4$+PW-S8 settings are paired within each allocation method; the accounting convention is the same as in Table~\ref{tab:task-lengths}.}
\label{tab:budget-task-lengths}
\begin{tabular*}{\textwidth}{@{\extracolsep{\fill}}llrrrrrrrr@{}}
\toprule
Model & Method & AIME & HMMT & GPQA & HE+ & MATH & IFB & LCB & Mean\\
\midrule
4B & STEP@$5$ & 69.58 & 80.97 & 41.15 & 1.27 & 23.02 & 18.96 & 25.77 & 37.25\\
 & STEP@$4$+PW-S8 & 70.31 & 81.62 & 34.76 & 1.44 & 19.68 & 12.85 & 25.38 & 35.15\\
\cmidrule(lr){2-10}
 & GGA@$5$ & 60.99 & 73.37 & 33.32 & 1.27 & 17.41 & 14.99 & 24.56 & 32.27\\
 & GGA@$4$+PW-S8 & 54.13 & 67.48 & 30.22 & 1.26 & 15.06 & 11.25 & 24.10 & 29.07\\
\midrule
9B & STEP@$5$ & 62.23 & 76.95 & 40.44 & 1.05 & 19.44 & 14.31 & 30.79 & 35.03\\
 & STEP@$4$+PW-S8 & 60.25 & 71.16 & 35.54 & 1.12 & 16.20 & 13.97 & 28.47 & 32.39\\
\cmidrule(lr){2-10}
 & GGA@$5$ & 46.49 & 60.65 & 29.06 & 1.03 & 13.18 & 13.14 & 23.91 & 26.78\\
 & GGA@$4$+PW-S8 & 43.23 & 53.68 & 24.53 & 1.05 & 11.78 & 10.71 & 22.24 & 23.89\\
\bottomrule
\end{tabular*}
\end{table}

\end{document}